\documentclass[11pt]{article}
\usepackage[final]{acl}
\usepackage{times}
\usepackage{latexsym}
\usepackage[T1]{fontenc}
\usepackage[utf8]{inputenc}
\usepackage{microtype}
\usepackage{graphicx}

\usepackage{booktabs}
\usepackage{amsmath}
\usepackage{amssymb}
\usepackage{array}
\usepackage{placeins}
\usepackage{makecell}

\newcommand{\ld}{\mathrm{LD}}
\newcommand{\margin}{\mathrm{margin}_3}

\title{Copying Before Suppression: What Drives a Below-Chance Dip During Language Model Training?}

\author{
  Tejas Dahiya \\
  University of Chicago \\
  \texttt{tdahiya@uchicago.edu}
  \And
  Cole Blondin \\
  Independent Researcher \\
  \texttt{ctblondin@gmail.com}
}

\begin{document}
\maketitle

\begin{abstract}
Mechanistic interpretability usually studies fully trained models, yet the computations that drive a behaviour can change while the model is still learning the task. On the Indirect Object Identification task, a model reads a sentence such as ``When Mary and John went to the store, John gave a drink to'' and should continue with Mary, the name mentioned once, rather than John, the name mentioned twice. Pythia models pass through an early training window in which they prefer the repeated name, so accuracy in a choice between the two names falls below one half while language-model loss on a fixed text sample keeps decreasing across the same window. The window reflects a temporary imbalance between two computations. We identify one cause of the wrong preference by selecting a set of attention heads that write the repeated name, on prompts separate from those used for causal evaluation, keeping that selection fixed, and then replacing each head’s final-token output with its average output on a separate set of non-repeated-name prompts. This improves the correct-minus-repeated logit difference in a separately trained 160M model and in the official 160M, 410M, and 1B models. At 160M, the head that lowers the repeated name in the mature model shows little of its mature behaviour at this point. It directs less than one percent of its attention to the repeated mention, and its output makes almost no direct contribution to lowering that name’s logit. Both properties grow over the interval in which behaviour recovers. Across the 160M, 410M, and 1B models, transplanting the corresponding head’s mature parameters into the early checkpoint recovers 35 to 68 percent of the total improvement in the correct-minus-repeated logit difference seen by the end of training. Related early-to-late reversals appear at further Pythia scales, in two independently trained GPT-2 models, and in OLMo. A mature circuit can therefore conceal a transient causal configuration that shaped behaviour earlier in training.
\end{abstract}

\section{Introduction}

Much work in mechanistic interpretability seeks to reverse engineer circuits, subgraphs of neural networks that implement interpretable model behaviours \citep{olah2020zoom}. One of the most thoroughly studied examples is indirect object identification (IOI), in which a model completes a sentence such as ``When Mary and John went to the store, John gave a drink to'' with the indirect object ``Mary.'' In GPT-2 Small, this behaviour is implemented by a circuit containing attention heads with distinct roles, including moving candidate names toward the output and suppressing the repeated subject \citep{wang2023ioi}. Mechanistic analyses have identified structured computations in other settings as well, including induction heads involved in in-context learning \citep{olsson2022induction} and a circuit for numerical greater-than comparisons in GPT-2 Small \citep{hanna2023greaterthan}.

These studies show how trained models implement particular behaviours, but the mechanism visible at the end of training need not reveal how that mechanism developed. A growing line of work therefore studies model internals across training. Prior work has tracked the development of linguistic representations \citep{saphra2019svcca}, the emergence of induction heads \citep{olsson2022induction}, the gradual internal changes underlying grokking \citep{nanda2023progress}, and the evolution of task circuits across training and model scale \citep{tigges2024llmcircuits}. Together, these results motivate studying not only which computations are present in a trained model, but also when they emerge and how their causal roles change during training.

In this work, we study this question in IOI using the Pythia model family \citep{biderman2023pythia}. Across the primary Pythia scales and independently trained variants, forced-choice accuracy between the correct indirect object (IO) and the repeated subject (S) falls substantially below chance early in training before recovering. In the released Pythia models, the floor occurs within the first 3000 training steps, with performance recovering soon afterwards. For Pythia 160M, this deterioration occurs while language-model loss on a fixed sample of Pile text continues to decrease. Related early-to-late causal reversals also appear at additional model scales, in independently trained GPT-2 models, and in OLMo, although we do not assume that these settings share the same underlying circuit.

This behaviour is difficult to explain with a simple picture in which task-relevant components strengthen together and performance improves monotonically. We instead test whether the below-chance window reflects a temporary imbalance between computations that develop at different times.

Our causal experiments support this account. At the floor, a set of attention heads makes a prompt-specific contribution that increases the model's preference for the repeated subject, and changing the identity supplied at their earlier source position redirects which name they support. Later in training, a suppressing head selected at maturity becomes strongly engaged and makes a substantial causal contribution to the correct preference. Transplanting its mature parameters into the floor model partially moves behaviour toward the mature state. We also localise an early-to-late reversal in causal role to the residual-stream state at the second subject mention. Together, these results show that a mature circuit can conceal a transient causal configuration present earlier in training, and that aggregate measures such as language-model loss can mask non-monotonic dynamics in particular behaviours.

\section{Setup}
\label{sec:setup}
\paragraph{Models and checkpoints.}
Our main analyses use the official Pythia 160M, 410M, and 1B models \citep{biderman2023pythia}. Each model has 154 released checkpoints. These include the initialisation checkpoint, 10 early checkpoints spaced logarithmically through step 512, and checkpoints every 1000 steps from step 1000 to step 143000. For our main analyses, we use the 1000-step checkpoints around the below-chance window.

We also analyse a separately trained Pythia 160M model with the same architecture and tokenizer. Around the below-chance window, this model has checkpoints every 50 steps, which lets us locate the floor and recovery more precisely. We use larger Pythia models, two Stanford CRFM GPT-2 training runs \citep{karamcheti2021mistral}, and OLMo \citep{groeneveld2024olmo} to test whether the early-to-mature reversal also appears at other model sizes and in other model families. We use PolyPythia \citep{vanderwal2025polypythias} for a different purpose. Its independently varied training runs let us test whether the below-chance window and the floor effect recur when the random seed, data order, or parameter initialisation changes.

For each training trajectory, we call the checkpoint with the lowest forced-choice accuracy the \emph{floor}. We call the last available checkpoint the \emph{mature checkpoint}. The floor occurs at step 2000 for the official Pythia 160M and 410M models, step 1000 for Pythia 1B, and step 2850 for the separately trained 160M model. We define the \emph{recovery checkpoint} as the first checkpoint after the floor at which forced-choice accuracy exceeds chance again, meaning accuracy is above $50\%$. The \emph{recovery interval} runs from the floor through the recovery checkpoint.

\paragraph{Task and notation.} We follow the standard IOI notation. Each prompt contains two single-token first names. The correct continuation, the indirect object $IO$, appears once. The other, the subject $S$, appears twice. We refer to the first and second mentions of the repeated name as $S1$ and $S2$, and we refer to the final sequence position, at which the model should predict IO, as $END$. For every sentence template, we permute the orders of IO and S1, so the position of a name alone does not reveal the correct answer. Figure~\ref{fig:task} in Appendix~\ref{app:protocols} illustrates this notation.

\paragraph{Evaluation.} Our primary comparison asks which of the two task-relevant names receives the higher logit at END. We define the \emph{logit difference} as
\begin{equation}
\ld = \ell(IO) - \ell(S),
\end{equation}
where $\ell(\cdot)$ is the logit at END. A positive value means that the model prefers the correct once-mentioned name, while a negative value means that it prefers the repeated name. Forced-choice accuracy is the fraction of prompts for which $\ld > 0$, whereas the reported mean logit difference averages $\ld$ across prompts. Accuracy therefore measures how often the correct name wins, while mean logit difference measures the average magnitude and direction of the preference. 

\begin{figure*}[t]
\centering
\includegraphics[width=0.78\textwidth]{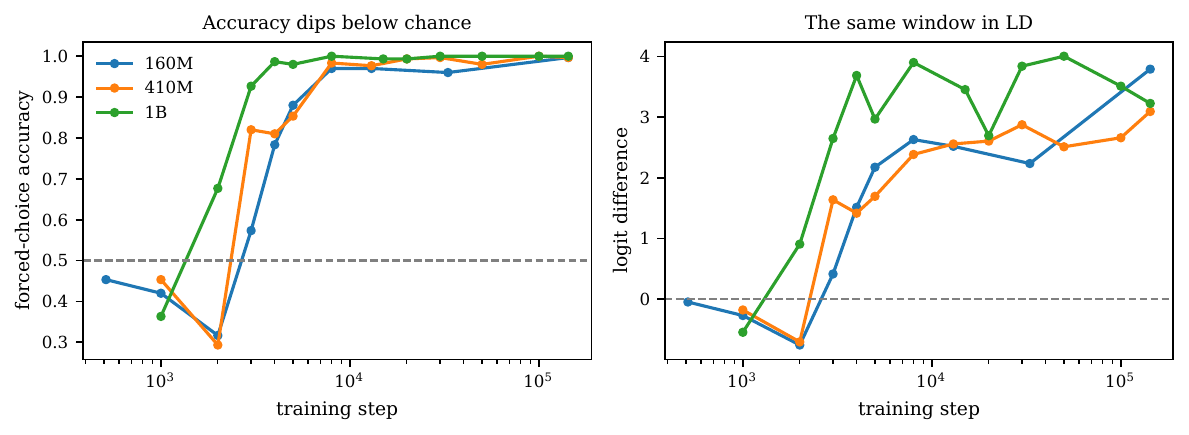}
\caption{The below-chance window. Panel A shows forced-choice accuracy at three model sizes falling below the chance level of one half and then recovering. Panel B shows the mean logit difference over the same interval. Panel A counts the fraction of prompts with positive logit difference, whereas Panel B averages its signed magnitude. The full-vocabulary output at the floor is shown in Figure~\ref{fig:appfullvocab}.}
\label{fig:phenomenon}
\end{figure*}

\section{A Below-Chance Window}
\label{sec:phenomenon}

\subsection{The behavioural reversal}
For all three official Pythia models, the same non-monotonic pattern appears during training. Forced-choice accuracy begins near chance, falls below one half to an early minimum, and later rises to near ceiling. The logit difference follows the same reversal, becoming negative during the window in which forced-choice accuracy is below chance and positive again after recovery (Figure~\ref{fig:phenomenon}). At the 160M floor, S receives mean probability $0.0238$, compared with $0.0118$ for IO, yielding a forced-choice accuracy of $0.317$. The mean vocabulary ranks are $14.9$ and $29.4$, respectively.

The floor does not coincide with a deterioration in next-token prediction. On a fixed sample of Pile text drawn from the same corpus that Pythia was pretrained on, language-model loss for Pythia 160M falls monotonically across the checkpoints surrounding the floor, from $3.73$ at step 1000 to $2.68$ at step 10000. This aggregate performance improvement masks task-specific performance degradation for IOI. Additional evaluations in Appendix~\ref{app:tasks} show that other structured tasks follow different trajectories over the same period of training.

\subsection{The window is robust across training runs and prompts}
The below-chance window is not specific to one training run. We examine nine additional 160M Pythia runs from the PolyPythia collection, in which either the random seed, the order of the training data, or the parameter initialisations are changed \citep{vanderwal2025polypythias}. Forced-choice accuracy falls below chance on every individual run, with mean floor accuracy of $27.7$ percent.

It also does not depend on the particular prompts. For Pythia 160M, using new names and new sentence templates gives a floor accuracy of $34.2$ percent. Using the new templates with the original names gives $30.0$ percent. Appendix~\ref{app:prompt_robustness} lists the names and templates used for these evaluations. Finally, we choose the floor checkpoint using one set of names and evaluate it using a separate set of names. This selects the same checkpoint, showing that the observed minimum is not an artefact of choosing and testing the checkpoint on the same examples.

These checks show that the below-chance window is a repeatable feature of the training trajectory rather than a peculiarity of one run or one evaluation set. Further controls in Appendix~\ref{app:window} show that the causal reversal at the second mention survives hand-written varied-syntax prompts, and that the behavioural window persists when candidate names span multiple tokens and are scored by full-name log probability.

\subsection{The causal role of the second subject mention reverses over training}
\label{sec:s2}

Having established the behavioural window, we next ask where in the prompt the relevant internal representation has a causal role that changes across training. We intervene separately at the first subject mention $S1$, the second subject mention $S2$, the indirect object $IO$, and a structural control position given by the token immediately after $S1$.

For the main $S2$ intervention, we use residual-stream activation patching. We call the evaluated prompt the \emph{destination} and the prompt supplying the replacement activation the \emph{source}. For example, the destination may contain ``Mary and John ... John gave a drink to,'' while the source contains ``Mary and John ... Alice gave a drink to.'' The source preserves the sentence structure and the once-mentioned name while replacing the second occurrence of the repeated name. It is used only to supply the activation at $S2$. Its own next-token prediction is not evaluated, and no correctness label for the source prompt is used. For each layer $\ell$ in a fixed layer band $B$, we replace the destination residual-stream activation at $S2$ with the corresponding source activation,
\[
h_{\ell,S2}^{\mathrm{dest}}
\leftarrow
h_{\ell,S2}^{\mathrm{source}}.
\]
The destination tokens and scoring remain unchanged, and the remainder of the forward pass proceeds normally.

We use layers 3 to 5 at 160M, 6 to 10 at 410M, and 4 to 8 at 1B, keeping each band fixed across checkpoints rather than selecting it separately at the floor and maturity. In a separate sensitivity analysis, we repeat the intervention across four non-overlapping layer bands in the separately trained 160M model at step 2000. The effect is positive in every band, although it is smaller in the final band. This shows that the positive early effect in this model is not restricted to the main layer band, while its magnitude still varies with depth. Appendix~\ref{app:bands} reports this analysis, while Appendix~\ref{app:s2} gives the bands used at larger scales.

We also perform a second $S2$ intervention that does not use a source prompt. At each patched layer, we replace the destination activation with an independently sampled Gaussian vector that is rescaled to match the norm of the clean activation for that prompt, layer, and position. This disrupts the $S2$ activation without importing information from another prompt.

At the 160M floor, the unmodified model has a mean logit difference of $-0.753$, indicating an average preference for the repeated name. Matched $S2$ activation patching raises the mean to $+0.200$, a change of $+0.953$. The intervention therefore reverses the direction of the model's average preference. Random $S2$ replacement produces almost the same result, raising the mean logit difference to $+0.167$, a change of $+0.920$.

The pattern reverses at maturity. The unmodified mature model has a mean logit difference of $+3.790$, indicating a strong average preference for the correct name. Matched and random $S2$ replacement reduce this value to $-0.333$ and $-0.336$, corresponding to changes of $-4.123$ and $-4.126$. Thus, disrupting the $S2$ state improves the relative preference at the floor but harms it after training. The close agreement between the matched and random replacements indicates that the reversal is driven by disrupting the $S2$ state rather than by transferring a particular alternative name from the source prompt.

\begin{figure}[t]
\centering
\includegraphics[width=\columnwidth]{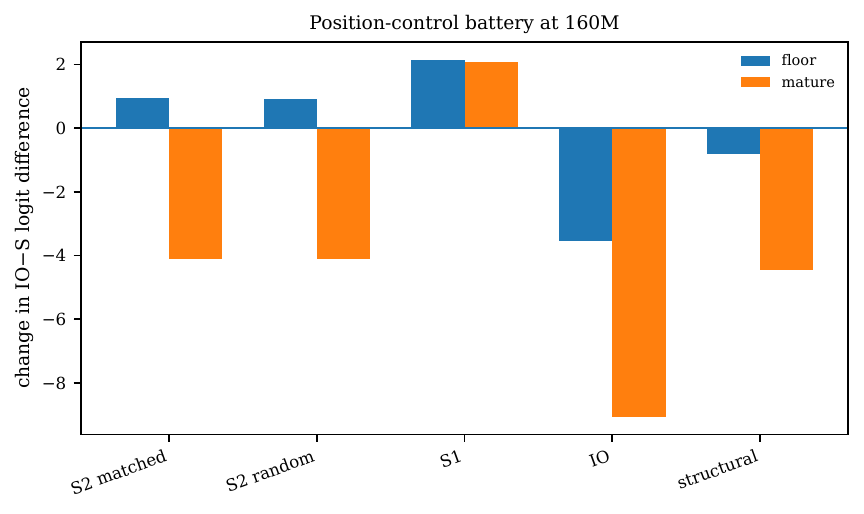}
\caption{Position controls at Pythia 160M. Replacing the residual-stream state at the second subject mention reverses sign between the floor and maturity for both the matched and random replacements. The tested interventions at the first subject mention, indirect object, and structural control position do not reverse sign.}
\label{fig:position_controls}
\end{figure}

We next apply the same random replacement at the other positions. At $S1$, the effect remains positive, changing from $+2.152$ at the floor to $+2.079$ at maturity. At $IO$, it remains negative, changing from $-3.536$ to $-9.063$. At the structural control position, it also remains negative, changing from $-0.815$ to $-4.450$. These positions can have large causal effects, but none changes sign between the floor and maturity (Figure~\ref{fig:position_controls}).

This experiment localises the reversal to the causal role of the $S2$ activation. Because activation patching replaces the entire residual representation at $S2$, it cannot tell us whether the harmful information reflects repetition, lexical identity, a particular attention-head contribution, or another feature. Section~\ref{sec:copying} next asks which component causes the repeated-name preference at the floor.

\FloatBarrier
\section{Writing the Repeated Name Is Causally Harmful}
\label{sec:copying}

\subsection{Testing the harmful contribution without a donor name}
\label{sec:m1design}

We first test whether, at the floor, a fixed set of attention heads makes a prompt-specific contribution that causally increases the model's preference for the repeated name. Section~\ref{sec:whycopying} then tests whether this harmful contribution specifically tracks the identity of the repeated subject supplied at its earlier mention.

We identify candidate heads using prompts that are separate from the later causal evaluation. On a selection split, we replace every attention head individually and retain heads whose replacement improves both the correct name's probability and the three-name margin, defined as the correct-name logit minus the larger of the repeated-name and unrelated-name logits. This leaves 21 of 144 heads in the official 160M model, 12 of 144 in the separately trained 160M model, 17 of 384 at 410M, and 22 of 128 at 1B. We then rank the surviving heads by their replacement effect on a separate validation split and freeze the resulting set before evaluating any reported effect on held-out prompts. Appendix~\ref{app:design} gives the full three-stage selection procedure.

This selection identifies heads whose prompt-specific contribution is harmful under the replacement intervention. It does not establish that they copy the repeated name's identity. We test that stronger claim separately in Section~\ref{sec:whycopying}.

A simple causal test would replace each selected head's output with the output from another prompt. That design has an important confound. The donor prompt can itself carry a particular name, so an apparent improvement could result from inserting a competing identity rather than removing the harmful prompt-specific contribution of the selected heads.

We therefore use a control-mean replacement. After the head set has been frozen, the remaining names and sentence templates are divided into two disjoint groups. One group is used only to construct the replacement vectors, and the other is used only for evaluation. The two groups share neither names nor templates.

For the control-mean group, we use versions of the prompts in which the two relevant name positions contain different names rather than the same repeated name. Both name orders are included. For each frozen head, we average its output vector at END over all of these prompts, producing one fixed vector per head. Because this vector is averaged across many name pairs and both orders, it is not tied to a particular donor prompt or donor name.

We then evaluate on the separate held-out group, using prompts in which one name is repeated as in the original task. At END, the prompt-specific output of every frozen head is replaced by that head's fixed control mean, after which the rest of the forward pass proceeds normally. The intervention therefore asks whether removing the selected heads' prompt-specific contribution improves the model's preference without inserting a particular competing identity. Appendix~\ref{app:copying} sets out this separation in full.

We measure the effect in two ways. The first is the change in the two-name logit difference $\ld$ defined in Section~\ref{sec:setup}. The second is a stricter \emph{three-name margin}. Let $ALT$ be a third name that does not appear in the prompt. We define
\begin{equation}
\margin = \ell(IO) - \max\{\ell(S),\, \ell(ALT)\},
\label{eq:three-name-margin}
\end{equation}
where $\ell(\cdot)$ is the final-token logit. An increase in $\ld$ can occur simply because the repeated name falls, even if another unrelated name becomes more competitive. The three-name margin guards against this by requiring the correct name to improve relative to both the repeated name and the unrelated alternative.

We report 95 percent intervals from template-to-item nested resampling. This procedure resamples sentence templates before resampling items within each selected template, so prompts sharing a template remain grouped. Appendix~\ref{app:copying} gives the complete procedure and the corresponding item-clustered results.

\begin{figure}[htbp]
\centering
\includegraphics[width=\columnwidth]{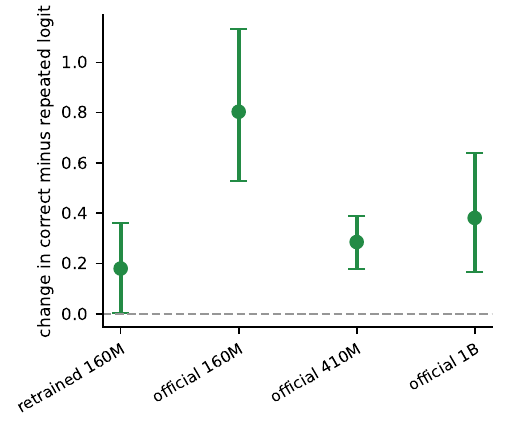}
\caption{Writing the repeated name is causally harmful at the floor. Replacing each frozen head's prompt-specific output at the final token with its control mean improves the correct minus repeated logit difference at all four evaluated floors. Bars are template-to-item nested intervals. The three-name margin is shown in Figure~\ref{fig:appthreename}.}
\label{fig:copying}
\end{figure}

\subsection{The replacement improves behaviour at every floor}
\label{sec:m1result}
Replacing the frozen heads' prompt-specific outputs with their control means improves the $IO$ versus $S$ logit difference at all four evaluated floors. The changes are $+0.180$ for the separately trained 160M model, $+0.804$ for official 160M, $+0.285$ for 410M, and $+0.381$ for 1B, with every template-aware 95 percent interval above zero.

The stricter three-name margin has the same positive direction in all four settings, although its template-aware interval lies above zero only for official 160M and 410M. The uniform conclusion is therefore that the selected heads causally worsen the model's preference for the correct name relative to the repeated name. Full intervals and both uncertainty analyses are reported in Appendix~\ref{app:copying}.

\subsection{Evidence that the selected heads copy name identity}
\label{sec:whycopying}
The experiment above establishes that their prompt-specific outputs worsen the correct-versus-repeated preference and that replacing them improves behaviour. It does not by itself show that the heads carry the identity of the name they read. A head could instead correlate with the repeated-name logit for some less specific reason.

We test name-identity transfer with a source-specific intervention. For a target prompt in which the same name appears at both subject positions, we construct a donor prompt that leaves the identity at $S2$ unchanged but substitutes the other name at $S1$. We then replace only the frozen heads' value vectors at $S1$ with those from the donor. The target prompt, the identity at $S2$, and all downstream computation are otherwise left unchanged.

If these heads merely happened to raise the repeated-name logit, changing the identity supplied at their source would not require their downstream contribution to follow the substituted name. Instead, the contribution does follow the substituted identity. At every floor, changing the identity supplied at $S1$ moves the ensemble's contribution away from the original repeated name and toward the substituted name. The resulting change in the three-name margin is $+0.412$ at the retrained 160M model, $+0.972$ at official 160M, $+0.596$ at 410M, and $+0.212$ at 1B. Both the item-clustered and the template-to-item nested intervals lie above zero in all four settings. Appendix~\ref{app:copying} gives both sets of intervals.

The selected heads therefore do more than correlate with a high repeated-name logit. Their contribution follows the identity supplied at the source position, which is why we describe them as copying heads. Appendix~\ref{app:alternatives} tests several simpler alternative explanations. We next ask what changes during training as the model recovers.

\section{Suppression Develops Later}
\label{sec:suppression}
Section~\ref{sec:copying} showed that repeated-name writing is already harmful at the floor. We now ask what changes as the model recovers. In the mature model, some attention heads help by lowering the repeated name. We therefore identify the head that makes the largest such contribution at maturity and trace that same head backward through training.

\subsection{Finding the mature suppressing head}
At the mature checkpoint we test each attention head one at a time. We replace the vector produced by that head with its average over the prompts and token positions of each sentence-template block, leaving the rest of the model unchanged, and measure how much this weakens the preference for the correct name over the repeated name. The head whose replacement produces the largest drop is selected as the mature suppressing head.

This procedure selects L8H9 at 160M, L12H12 at 410M, and L11H0 at 1B. We then examine what these heads do in the unmodified mature model. All three behave as expected of a head that helps suppress the repeated name. From the final prediction position END they attend strongly to the second occurrence of the repeated name, with mean attention $0.934$ at 160M, $0.336$ at 410M, and $0.527$ at 1B. Their outputs also point in a direction that lowers the repeated name's logit, with direct output projections of $-6.22$, $-2.10$, and $-7.27$. These measurements characterise the selected heads rather than confirming the selection, since the replacement effect used to choose them is a causal quantity and the two descriptive quantities are reported alongside it.

\subsection{The mature suppressing head is weakly engaged at the floor}
We next trace each mature-selected head backward through training without choosing a new head at earlier checkpoints. This lets us ask when the computation that matters at maturity first becomes active.

At the 160M floor the mature suppressing head places only $0.0088$ of its attention from END on the second mention and its direct output projection on the repeated name is only $-0.034$. One thousand steps later, as behaviour begins to recover, its attention rises to $0.849$ and its projection grows to $-1.338$. By maturity these reach $0.934$ and $-6.22$.

The same contrast appears at the other primary sizes. At the 410M floor the head places $0.062$ of its attention on the second mention with projection $-0.005$, rising to $0.722$ and $-0.246$ at the next measured checkpoint. At 1B, where the trajectory is sparser, the quantities move from $0.014$ and $-0.025$ at the floor to $0.527$ and $-7.27$ at maturity.

The trajectory is therefore not identical across sizes, and the 1B checkpoints do not resolve its detailed shape. What is consistent is the contrast between the floor and later training. The computation that strongly suppresses the repeated name at maturity is barely engaged when the model is most wrong, and becomes substantial around recovery.

\subsection{The suppressing head contributes substantially at maturity}
Using the same replacement procedure, replacing the selected head lowers the mature logit difference by $2.66$ at 160M, $1.76$ at 410M, and $1.59$ at 1B. The model does not lose the behaviour entirely: forced-choice accuracy remains $0.78$, $0.90$, and $0.99$. The head therefore contributes substantially to the strength of the mature preference without being solely responsible for mature behaviour.

Replacing the selected head substantially weakens the mature preference while forced-choice accuracy remains above chance at all three sizes. The head is therefore an important part of the recovered computation, while other components also contribute.

\subsection{Mature suppressor parameters partially rescue the floor}
The results above show that the suppressing computation is weak at the floor and strong later, but not whether the changes this head learns are themselves enough to improve the early model.

We therefore take the parameters learned by the mature suppressing head and place them into the same head at the floor checkpoint, replacing its query, key, value, and output parameters with their mature values. Everything else in the floor model is unchanged, and the early model then runs normally using its own activations. The intervention changes what the head has learned rather than inserting an activation produced by the mature model.

The result is a substantial partial rescue at all three sizes. At 160M the logit difference moves from $-0.753$ at the floor to $+0.849$. At 410M it moves from $-0.702$ to $+1.876$, and at 1B from $-0.543$ to $+0.984$. These changes close $35.3$, $67.9$, and $40.5$ percent of the distance between the floor and mature logit differences, and forced-choice accuracy rises to $0.687$, $0.893$, and $0.810$.

The rescue is large but incomplete. The mature suppressing head contains learned changes sufficient to reverse the early preference on many prompts, yet the recovered model cannot be reduced to this head alone. Appendix~\ref{app:suppq} gives additional controls, including a matched activation experiment and parameter transplants into unrelated heads.

At the floor, repeated-name writing is already causally harmful while the computation that later suppresses that name is almost absent. During recovery suppression becomes strongly engaged, and giving the early model the mature suppressor parameters moves its behaviour substantially toward the mature state. The below-chance window is therefore best understood as a temporary imbalance between two computations that develop at different times, rather than as the mature circuit operating more weakly.

\begin{figure*}[tbp]
\centering
\includegraphics[width=0.98\textwidth]{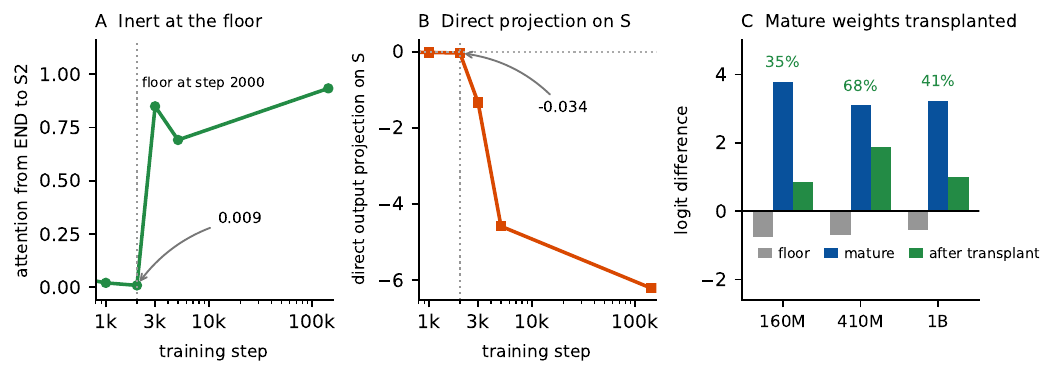}
\caption{The suppressing head develops during recovery. Panel A shows that at the 160M floor the head places under one percent of its final-token attention on the repeated mention, rising sharply as behaviour recovers. Panel B shows that the head's direct output projection on the repeated name is close to zero at the floor and becomes strongly negative later. Panel C shows that transplanting the mature head's parameters into the early checkpoint moves the logit difference a substantial fraction of the way to the mature value at all three sizes. The effect of replacing this head's output in the mature model is shown in Figure~\ref{fig:appablation}.}
\label{fig:suppression}
\end{figure*}

\section{Scope Beyond the Primary Models}
\label{sec:scope}
The detailed copying and suppression account above is established for Pythia 160M, 410M, and 1B. We next ask whether the same broad developmental pattern appears at other scales, in other model families, and on other tasks. These experiments test the pattern rather than whether every setting contains the same heads.

\subsection{Across scale and model family}
Early sensitivity to repetition extends across Pythia scale, though not every size is tested under the same intervention. Using first-feed-forward-block replacement, that is the layer-0 MLP, we replace its output at 70M, 160M, 410M, 1B, and 1.4B with the corresponding output from a matched prompt in which the name is not repeated. This improves the floor logit difference at every size, with all five intervals above zero, while the unmodified floor logit difference is negative in every case. At 2.8B, replacing the second occurrence of the repeated name with a different name in the input improves the floor logit difference by $+0.456$, while the same edit changes it by $-4.076$ at maturity. Matched second-mention state replacement reverses in the same way at 6.9B and 12B, changing the logit difference by $+0.837$ and $+0.434$ at their floors and by $-4.118$ and $-3.955$ at maturity. Resampling the outputs of the early feed-forward blocks separately raises floor accuracy from $0.297$ to $0.603$. Repetition-sensitive information is therefore harmful early across a wider range of Pythia sizes. At 2.8B the input-edit effect additionally reverses by maturity.

The same ordering appears in two independently trained GPT-2 models released by the Stanford CRFM Mistral project \citep{karamcheti2021mistral}. At step 1500, replacing the first feed-forward block output with its non-repeated counterpart improves the logit difference by $+1.123$ and $+1.472$ in the two runs. At the same point, replacing the head that later becomes the mature suppressor changes the logit difference by approximately zero. At maturity the first intervention instead changes the logit difference by $-4.081$ and $-4.607$, while replacing the mature suppressing heads changes it by $-1.928$ and $-1.506$. The suppressing head differs between the runs, as do the detailed recovery trajectories. What repeats is the order in which the two computations become important.

OLMo-1B \citep{groeneveld2024olmo} provides a third family. At its early checkpoint, forced-choice accuracy is $0.437$ and the logit difference is $-0.220$. Replacing the residual-stream activation at the second occurrence of the repeated name with the corresponding activation from a matched non-repeated prompt improves the logit difference by $+0.0448$, with interval $[+0.0349, +0.0549]$. At maturity the same intervention changes it by $-5.137$, with interval $[-5.257, -5.018]$. The causal role of that activation therefore also reverses over training outside the Pythia and GPT-2 families. Appendix~\ref{app:scale} gives the full protocol for each setting.

\subsection{Across tasks}
A related reversal appears on Greater-Than \citep{hanna2023greaterthan}. For a prompt such as ``The war lasted from the year 1723 to the year 17'', the model should prefer a completion greater than 23. We construct a matched prompt with a different starting year and replace the residual-stream activation at the token carrying the starting year's final two digits with the corresponding activation from that matched prompt. At step 1000 this improves the task score by $+0.082$, $+0.097$, and $+0.319$ at 160M, 410M, and 1B. At maturity the same intervention changes the score by $-0.737$, $-0.667$, and $-0.732$. The 410M model is already above chance at the early checkpoint, with accuracy $0.553$, so that setting shows a reversal in what the state contributes rather than another below-chance behavioural window.

Taken together, these results support a recurring developmental pattern rather than a single circuit shared across models and tasks. Information that is harmful, or useful to disrupt, early in training can become useful later, while computations that matter at maturity can be weak during the earlier failure. The detailed account in which repeated-name copying is followed by the later development of suppression remains specific to the three primary Pythia models.

\section{Related Work}

Mechanistic interpretability increasingly examines how model computations develop during training. \citet{saphra2019svcca} track the emergence of linguistic representations, \citet{olsson2022induction} connect the formation of induction heads with in-context learning, and \citet{nanda2023progress} identify gradual internal changes underlying grokking. \citet{tigges2024llmcircuits} further show that broad algorithms and functional roles can remain stable across training and scale even when the individual components implementing them change. Together, these studies show that internal structure can develop gradually and need not be fully understood from the final checkpoint. We extend this developmental view to a temporary failure in which behaviour becomes worse than at both earlier and later checkpoints, and we investigate the causal computations that create and resolve that failure.

Our analysis also builds on prior work on the mature IOI circuit. \citet{wang2023ioi} identify attention heads that move candidate names toward the output and suppress the repeated subject in GPT-2 Small. \citet{mcdougall2024copysuppression} examine copy suppression more broadly and show how attention heads can reduce support for tokens promoted by earlier components. These studies establish the mature copying and suppression mechanisms that motivate our analysis. We examine how their causal balance changes during training and find that repeated-name writing is already harmful when behaviour reaches its floor, while the suppressing computation that later supports the correct answer is still weak.

Our experiments draw on causal mediation, interchange interventions, activation patching, and path patching \citep{vig2020causal,geiger2021causal,meng2022locating,conmy2023automated,goldowskydill2023localizing}. Prior work has shown that conclusions from these methods can depend on the replacement used and the quantity being measured \citep{zhang2024bestpractices,heimersheim2024activation}. We therefore use separate interventions for separate causal questions. Control-mean replacement tests whether selected head outputs are harmful, source-position value replacement tests whether their contribution follows a name’s identity, and parameter transplantation tests whether a computation learned later can improve the earlier model.

\section{Discussion}
Our results show that the mechanism visible at the end of training can give an incomplete picture of how a behaviour developed. At the floor, repeated-name copying is already harmful while the suppressing computation that matters at maturity is still weak. As training continues that balance changes. Intermediate checkpoints can therefore contain causal structure that is largely invisible from the final model alone.

More broadly, Section~\ref{sec:scope} shows related early-to-late reversals across additional model sizes, GPT-2, OLMo, and Greater-Than. We do not read this as evidence for one shared circuit across all settings. It suggests a broader developmental pattern in which information or computations that are harmful, or useful to disrupt, early in training can later become useful as other parts of the model develop. For mechanistic interpretability, this makes the timing of a computation's contribution as important as its role in the final model.

\section{Conclusion}
During the below-chance window, repeated-name copying is already causally harmful while the suppression that later supports correct behaviour is still weak. As the model recovers, suppression becomes important, and giving the early model the mature suppressor parameters partially moves its behaviour toward the mature state. These results show that the mechanisms governing behaviour during training can differ substantially from those visible at the final checkpoint, motivating mechanistic analysis of how circuits develop rather than only how they operate once training is complete.

\section*{Limitations}
Our strongest causal account is established for the naming task in the Pythia 160M, 410M, and 1B models. Additional Pythia scales, two independently trained GPT-2 models, OLMo, and Greater-Than show related developmental reversals, but we do not establish that they use the same copying and suppression mechanism. The broader evidence supports a recurring pattern across training rather than a universal circuit shared across models and tasks.

The interventions also answer different causal questions. Replacing the residual-stream state at the second name mention localises where information affecting the prediction is carried, but does not identify a duplication-specific feature. The copying claim instead rests on interventions on a selected set of attention heads, replacing their prompt-specific outputs and changing the identity supplied at the source position. These give stronger evidence for repeated-name copying, but the head-level account is limited to the primary models.

Our analysis explains the below-chance behaviour as an imbalance between harmful copying and weak suppression, but not why training develops copying first. Suppression may become useful only after the model develops a pathway that makes the repeated candidate available, or the order may be driven by the training data. Distinguishing these possibilities would require changing the data or intervening during training to change how copying develops, then measuring the effect on suppression.

Finally, our experiments focus on relatively small open-weight models and structured tasks. For the released Pythia runs, checkpoints around the behavioural transition are spaced 1000 steps apart, limiting how precisely we can localise when computations emerge or change causal role. This leaves open how often the same pattern appears in larger models, less structured behaviour, or settings with sparser checkpoints.

\bibliography{references}

\clearpage
\appendix
\section{Protocols and Manipulations}
\label{app:protocols}

\begin{figure*}[tp]
\centering
\includegraphics[width=\textwidth]{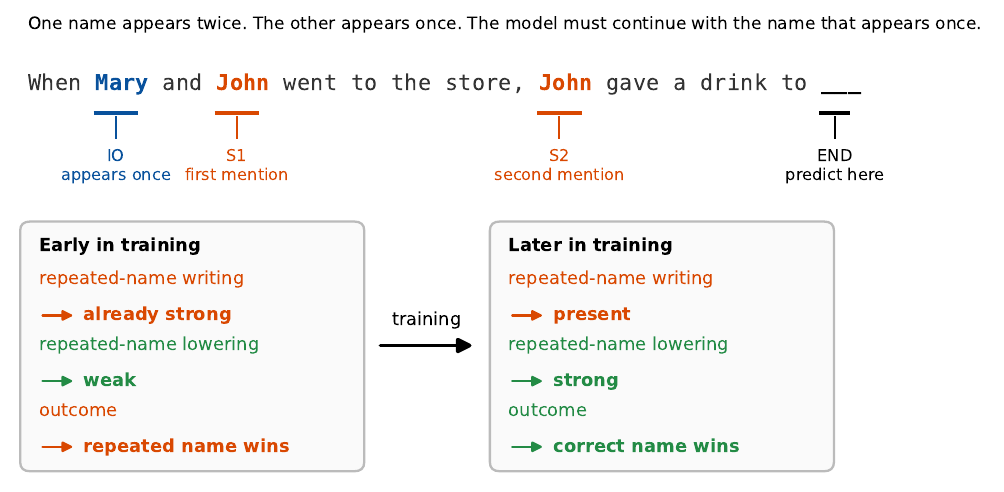}
\caption{The task and the account in one picture. The upper line shows a prompt with the indirect object mentioned once, the subject mentioned twice, and the position where the model must answer. The lower panels summarise what this paper establishes. Early in training the heads that copy the repeated subject are already causally effective while the head that later suppresses it is close to inert, and the model falls below chance. Later the suppressing computation grows and behaviour recovers.}
\label{fig:task}
\end{figure*}

\subsection{Prompt construction}
We follow the prompt construction of \citet{wang2023ioi}. Each prompt instantiates a
sentence template with two distinct first names that each correspond to a single token.
One name appears twice, at $S1$ and $S2$, while the other appears once and is the correct
continuation at END. Figure~\ref{fig:task} gives a concrete example.

For each template and name pair, we balance which name is repeated and which is the
correct once-mentioned name. This prevents either the identity or the earlier position of
a name from determining the answer. When an experiment requires separate prompts for selection,
replacement construction, and final evaluation, the relevant names
and sentence templates are partitioned before the intervention is evaluated. The
experiment-specific sections give the exact split used in each case.

\subsection{Models and checkpoints}
Our primary mechanistic analyses use the released Pythia 160M, 410M, and 1B models
\citep{biderman2023pythia}. We additionally use a separately trained 160M model with the
same architecture and tokenizer but more frequent checkpoints around the early transition.
This model helps resolve when the relevant computations change, but the main phenomenon
and the cross-scale conclusions do not depend on that run.

Additional Pythia sizes from 70M through 12B are used for the scale analyses described in
the corresponding appendices. Appendix~\ref{app:scale} separately gives the model and
checkpoint details for the GPT-2 and OLMo experiments.

For each run, the floor is the evaluated checkpoint with the lowest forced-choice
accuracy. It occurs at step 2000 for the released 160M and 410M models, step 1000 for 1B,
and step 2850 for the separately trained 160M model. We call the final checkpoint the mature checkpoint, and the first evaluated checkpoint after the floor at which forced-choice accuracy exceeds $50\%$ the recovery checkpoint.

\subsection{Intervention glossary}
\label{app:ops}

The experiments below modify different parts of the model and answer different causal questions. We therefore use a distinct name for each intervention and specify both the quantity being replaced and the source of its replacement.

\begin{figure*}[htbp]
\centering
\includegraphics[width=\textwidth]{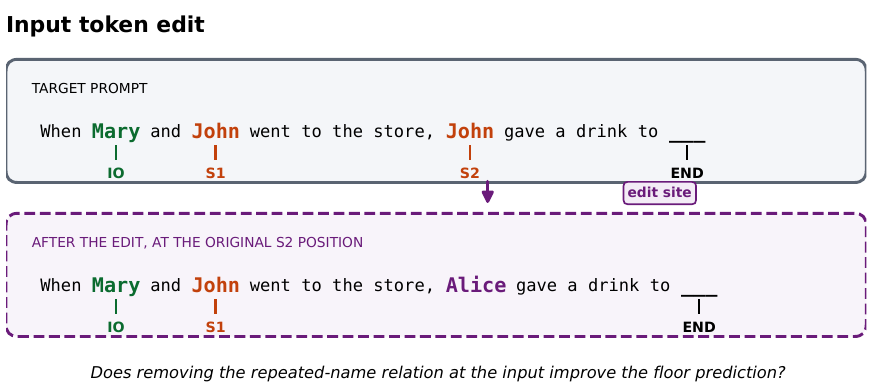}
\caption{Input token edit. The second occurrence of the repeated name is replaced in the token sequence itself, and the model then runs normally. Because repetition status and token identity change together, a positive effect is evidence that the model is sensitive to the repeated relation, not an isolation of repetition alone. This is the only manipulation in the paper that alters the evaluated prompt.}
\label{fig:atlasB}
\end{figure*}

\paragraph{Input token edit.} The token sequence itself is changed before the model runs.
For example, the second occurrence of the repeated name can be replaced by a third name.
This tests sensitivity to the repetition in the input, but does not identify where or how
the model uses that information. Figure~\ref{fig:atlasB} shows the edit on the canonical prompt, with the token that changes marked.

\paragraph{Residual-stream state replacement.} The residual-stream activation at one token
position is replaced across a specified band of layers. The replacement can come from a
matched prompt or from an independently sampled direction whose magnitude is matched to
the original activation. This localises where a causally important state is carried, but
changing the entire state does not identify which feature within it matters.

\paragraph{First feed-forward block activation replacement.} The output of the first
block's feed-forward module is replaced with the corresponding output from a matched
prompt in which the name is not repeated. The evaluated prompt itself is unchanged. This
tests whether repetition-sensitive information computed early in the network is harmful,
without treating the intervention as an input edit.

\paragraph{Control-mean head-output replacement.} For the copying experiment in
Section~\ref{sec:copying}, each frozen head's output at END is replaced with one fixed
vector computed by averaging that head's output over separate non-repeated control
prompts. Because the average spans many names and both name orders, it is not tied to any
particular donor identity. This tests whether the selected heads' prompt-specific
contribution is causally harmful.

\paragraph{Source-position value replacement.} To test whether that contribution actually
follows the copied name, we replace only the frozen heads' value vectors at $S1$ with
those from a prompt carrying a different name there. The target prompt, the identity at
$S2$, and downstream computation remain unchanged. If the downstream contribution follows
the substituted identity, the intervention provides direct evidence of name copying.

\paragraph{Passive attention split.} The total attention assigned to $S1$ and $S2$ is
preserved but divided equally between the two positions, while their value vectors remain
unchanged. This tests whether the observed attenuation requires a learned routing decision beyond the mechanical distribution of a fixed amount of attention across two positions.

\FloatBarrier

\paragraph{Mature-head mean replacement.} In the mature-suppressor experiments, the
selected head's activation is replaced by its mean over the prompts and token positions of each sentence-template block. This differs from the separate control-prompt mean used in the copying
experiment above. It measures how strongly the mature model depends on the selected head.

\paragraph{Activation graft.} An activation at a chosen site is replaced by the
corresponding activation produced by another checkpoint of the same model on the same
prompts. This asks whether the later activation itself is sufficient to change earlier
behaviour.

\paragraph{Parameter transplant.} The query, key, value, and output parameters of one
attention head are replaced with those of the same head at another checkpoint. The
receiving model then computes its own activations normally. Our main use copies the mature
suppressor's parameters into the floor model and asks whether what the head learned later
is sufficient to improve early behaviour.

\paragraph{Route isolation.} A specific route between two components is isolated while
alternative routes are held fixed by the intervention. This estimates how much of a
measured dependence travels through that particular route.

\subsection{Uncertainty and resampling}
We use several resampling procedures depending on the structure of the evaluation set. A
prompt bootstrap resamples individual prompts. An item-clustered bootstrap instead keeps
paired evaluations of the same item together before resampling, so closely related
observations are not treated as independent.

For the copying experiments, prompts also share sentence templates. We therefore report a
stricter template-to-item nested bootstrap that first resamples templates and then
resamples items within each selected template. The intervals quoted in the main text use
this procedure with ten thousand bootstrap draws.

Because the released settings contain only three template groups and the separately
trained setting contains four, these nested intervals rest on a small number of top-level
groups. We therefore report the item-clustered intervals alongside them rather than
treating either procedure as the sole uncertainty estimate.

\section{Behavioural Characterisation and Robustness}
\label{app:window}
This appendix gives additional evidence that the below-chance window is a genuine and
repeatable change in the model's preference between the two task-relevant names. We
examine the full output distribution, ordinary language-model loss, independently varied
training runs, and evaluations using prompts separated from those used elsewhere in the
analysis.

\subsection{Full-vocabulary behaviour at the floor}
Forced-choice accuracy compares only the correct name $IO$ and the repeated name $S$. We
therefore examine the complete next-token distribution to determine what the below-chance
result means outside that two-name comparison.

\begin{figure}[htbp]
\centering
\includegraphics[width=\columnwidth]{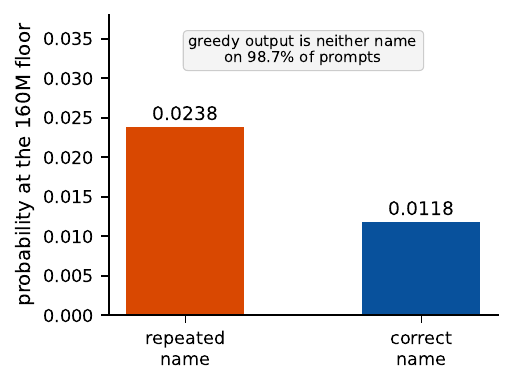}
\caption{Full-vocabulary output at the 160M floor. The repeated name receives about twice the probability of the correct name, while greedy decoding produces neither name on almost all prompts.}
\label{fig:appfullvocab}
\end{figure}

At the 160M floor, the repeated name receives mean probability $0.0238$, compared with
$0.0118$ for the correct name. Its mean vocabulary rank is also better, $14.9$ compared
with $29.4$ for the correct name. Greedy decoding, however, produces neither candidate on
$98.67$ percent of prompts. Figure~\ref{fig:appfullvocab} shows the probability comparison
at the floor.

\begin{figure}[htbp]
\centering
\includegraphics[width=\columnwidth]{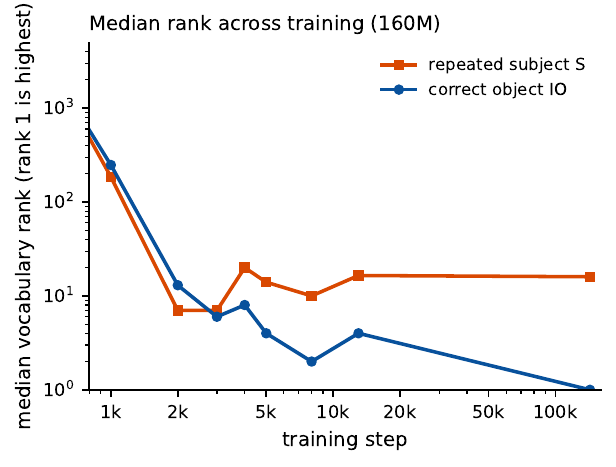}
\caption{Vocabulary rank across training at 160M. Median rank of the repeated and correct names across checkpoints, on a logarithmic scale, with rank 1 the highest-scoring token, so a lower rank indicates greater model preference. The repeated name ranks ahead around the floor and the correct name overtakes it during recovery.}
\label{fig:apprank}
\end{figure}

The below-chance window therefore describes a reversal in the relative preference between the two task-relevant names. At the floor, the repeated name is preferred to the correct name even though neither candidate is usually selected by greedy decoding. Figure~\ref{fig:apprank} follows the candidates' vocabulary ranks across
training. The repeated name is ranked ahead around the floor, after which the correct name overtakes it during recovery.

Logit difference is also not an artefact of using unnormalised logits. Because the correct
and repeated names are scored at the same next-token position in the same forward pass, the
normalising constant of the softmax is shared between them, so
$\log p(IO) - \log p(S) = z_{IO} - z_{S} = \ld$ exactly. We compute the two quantities
independently and verify this equality numerically at every evaluated checkpoint. As a
separate scale-free check, the repeated name also outranks the correct name in the full
vocabulary at each primary-model floor. Using rank 1 for the highest-scoring token, the
median repeated-name and correct-name ranks are 7 against 13 at 160M, 10 against 24.5 at
410M, and 57 against 98 at 1B, with floor accuracies of $0.317$, $0.293$, and $0.363$. The
below-chance ordering therefore appears in the normalised log-probability ratio and in
vocabulary rank alike, rather than depending on the numerical scale of the logits.

\subsection{Language-model loss continues to improve}
We next ask whether the failure on this task coincides with a broader deterioration in
next-token prediction. We evaluate the released Pythia 160M checkpoints on a fixed sample of Pile text drawn from the same corpus that Pythia was pretrained on.

Mean language-model loss decreases from $3.734$ at step 1000 to $3.140$, $2.964$, $2.801$,
and $2.677$ at steps 2000, 3000, 5000, and 10000. The decrease is monotonic across the
checkpoints spanning the floor and recovery.

The below-chance window therefore occurs while language-model loss on a fixed text sample continues to
improve. This does not imply that every other capability improves over the same interval,
but it rules out worsening language-model loss on this sample as an explanation for the
reversal. The IOI reversal therefore occurs while aggregate next-token loss on the sampled Pile text continues to improve. Appendix~\ref{app:tasks} compares the trajectories of other structured tasks over the same training period.

\begin{figure}[htbp]
\centering
\includegraphics[width=\columnwidth]{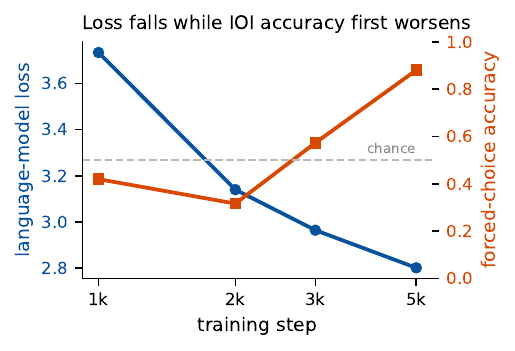}
\caption{Language-model loss and IOI forced-choice accuracy around the 160M floor. Loss decreases from steps 1000 to 5000, while IOI accuracy first falls and then recovers. The dashed line marks chance-level forced-choice accuracy.}
\label{fig:app_loss_accuracy}
\end{figure}

\begin{figure}[htbp]
\centering
\includegraphics[width=\columnwidth]{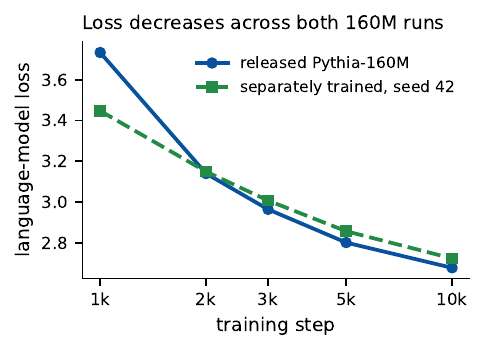}
\caption{Language-model loss on the same fixed text sample for the released Pythia 160M model and the separately trained 160M model. Loss decreases across the evaluated checkpoints in both runs.}
\label{fig:app_loss_runs}
\end{figure}

\subsection{Robustness across training runs}
The window also persists across independently varied training runs. We evaluate nine
PolyPythia 160M variants \citep{vanderwal2025polypythias}. Three change the random seed,
three change the order of the training data, and three change the parameter
initialisation. Across all nine runs, mean accuracy at the floor is $27.7$ percent.

\begin{figure}[htbp]
\centering
\includegraphics[width=\columnwidth]{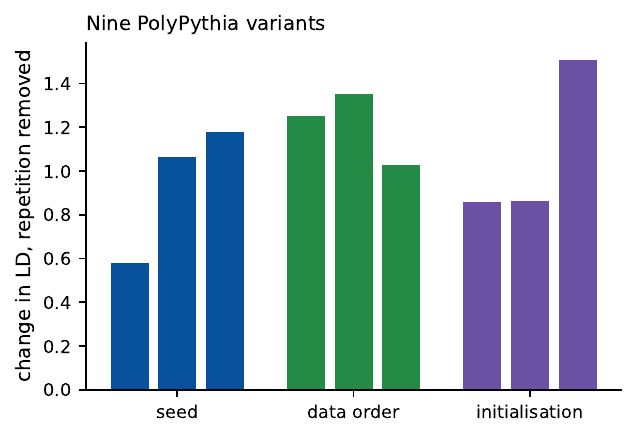}
\caption{Repetition sensitivity across nine PolyPythia 160M runs. It shows change in logit difference at each run's floor after replacing the second occurrence of the repeated name with a different name. All nine effects are positive. These runs are grouped by whether they vary the random seed, the data order, or the parameter initialisation.}
\label{fig:appnineseed}
\end{figure}

We additionally apply the input token edit defined in Appendix~\ref{app:ops}, replacing the
second occurrence of the repeated name with a different name. This improves the floor
logit difference in all nine runs, with mean change $+1.076$ and standard deviation
$0.267$. Figure~\ref{fig:appnineseed} shows the effects grouped by the aspect of training
that was varied.

The below-chance preference and its sensitivity to the repeated relation therefore survive changes in the random seed, data order, and parameter initialisation. This experiment establishes behavioural and input-level robustness across the nine runs. The head-level copying mechanism is evaluated separately in Appendix~\ref{app:copying}. Table~\ref{tab:polypythia_runs} reports the individual
results for all nine runs.

\begin{table}[t]
\centering
\footnotesize
\setlength{\tabcolsep}{3pt}
\begin{tabular}{@{}lrrrr@{}}
\toprule
Run & Floor acc. & Native LD & Edited LD & $\Delta$LD \\
\midrule
seed1        & 30.7\% & $-0.688$ & $-0.106$ & $+0.582$ \\
seed3        & 14.7\% & $-1.406$ & $-0.343$ & $+1.063$ \\
seed5        & 25.3\% & $-1.104$ & $+0.075$ & $+1.179$ \\
data-seed1   & 30.7\% & $-1.074$ & $+0.176$ & $+1.250$ \\
data-seed2   & 19.3\% & $-1.514$ & $-0.161$ & $+1.353$ \\
data-seed3   & 36.3\% & $-0.868$ & $+0.161$ & $+1.029$ \\
weight-seed1 & 29.0\% & $-0.927$ & $-0.069$ & $+0.858$ \\
weight-seed2 & 35.7\% & $-0.548$ & $+0.315$ & $+0.863$ \\
weight-seed3 & 27.3\% & $-1.116$ & $+0.392$ & $+1.508$ \\
\midrule
Mean & 27.7\% & & & $+1.076$ \\
\bottomrule
\end{tabular}
\caption{Results for the nine PolyPythia 160M variants. Floor accuracy is below the $50\%$ forced-choice baseline for every run. Native and edited LD are measured before and after removing the repeated relation.}
\label{tab:polypythia_runs}
\end{table}

\subsection{Robustness to new names and templates}
\label{app:prompt_robustness}

We test this robustness on Pythia 160M. With both new names and new templates, floor accuracy is $34.2$ percent. Using the new templates with the original names gives $30.0$ percent. In both cases accuracy later recovers to near ceiling.

\paragraph{Evaluation material.}
We use four new sentence templates that do not appear in the primary evaluation:

\begin{enumerate}
    \item ``After the meeting, \{S\} and \{IO\} grabbed coffee, and \{S\} handed the cup to''
    \item ``At the party, \{IO\} and \{S\} were chatting, and \{S\} passed a drink to''
    \item ``During the trip, \{S\} and \{IO\} shared a cab, and \{S\} gave directions to''
    \item ``In the lab, \{IO\} and \{S\} ran the test, and \{S\} showed the result to''
\end{enumerate}

For the new-template and new-name evaluation, the names are Daniel, Rachel, Kevin, Laura, Brian, Nicole, Steven, Megan, Jason, Emily, Aaron, and Diana. For the new-template-only evaluation, the names are John, Mary, Tom, Anna, Paul, and Lucy. We retain only names that correspond to a single token under the Pythia tokenizer.

There is a separate concern when defining the floor itself. Because the floor is chosen as the checkpoint with the lowest accuracy on this task, selecting and evaluating it on the same prompts could exaggerate the depth of the minimum. We therefore select the floor using one set of names and evaluate the resulting checkpoint on a separate name set. The independently selected floor is the same checkpoint.

These controls show that the observed window is not tied to the particular names or sentence templates in the primary evaluation, and is not an artefact of selecting the minimum on the same examples used to report it.

\subsection{Robustness to naturalised wording}
The preceding controls change the names and templates within the benchmark family. We
additionally test whether the phenomenon survives substantially less stereotyped wording,
using hand-written sentences with varied syntax. Each prompt still contains one candidate
name once and the competing subject twice, but the surrounding phrasing is written
independently rather than instantiated from the primary template set. We require both names
to be single tokens, verify the repeated-name structure after tokenisation, and locate the
second mention directly in each sequence rather than assuming a fixed offset. Thirty-four
prompts pass these checks. We evaluate the three primary Pythia sizes at their respective
floors and at maturity, applying the same matched residual-state intervention at the second
mention and the same model-specific layer bands used in the main analysis.

The reversal survives this change in wording. Floor accuracy is $0.029$, $0.147$, and
$0.147$ at 160M, 410M, and 1B respectively, so the behaviour is below chance on these
prompts at every size. At 160M the intervention improves the logit difference by $+0.882$
with interval $[0.626, 1.164]$, while at maturity the effect reverses to $-2.150$ with
$[-2.912, -1.391]$. The corresponding floor-to-mature effects are $+0.344$
$[0.165, 0.545]$ to $-3.899$ $[-4.330, -3.475]$ at 410M, and $+0.403$ $[0.292, 0.528]$ to
$-1.565$ $[-2.254, -0.911]$ at 1B. Every floor interval is strictly positive and every
mature interval strictly negative, so the below-chance behaviour and the causal reversal at
the second mention are not restricted to the rigid sentence templates of the primary
benchmark. These hand-written prompts establish robustness to substantially varied syntax, while broader natural-text generalisation remains a separate question.

\subsection{Robustness to multi-token names}
The primary experiments restrict the candidate names to single tokens so that the two
candidates can be compared at one next-token position. To test whether the behavioural dip
depends on that restriction, we construct prompts whose candidate names each contain
exactly two tokens. Candidate first-name and surname combinations are filtered through the
model tokenizer, leaving 178 valid two-token names, and 120 prompts are sampled across
three sentence frames. Rather than scoring only the first token, we teacher-force each
complete candidate continuation and sum the log probabilities of its two tokens. The
resulting score is the full-name log-probability margin between the correct and repeated
names.

Under this score the model remains strongly below chance at the 160M floor. At step 2000
the mean margin is $-2.255$ with interval $[-2.553, -1.954]$, and only $7.5$ percent of
prompts favour the correct complete name. At step 3000 the margin remains negative at
$-1.192$ with $[-1.550, -0.832]$ and accuracy $26.7$ percent, before recovering by maturity
to $+2.848$ with $[2.507, 3.197]$ and accuracy $93.3$ percent. The temporary preference
reversal therefore survives when candidate answers span multiple tokens and are scored as
complete names. This experiment is behavioural only. Extending the causal intervention at
the second mention to multi-token continuations would require a separate intervention and
scoring protocol, and we do not claim that the causal reversal has been tested here.

Together with the held-out-name, held-out-template, and naturalised-prompt controls above,
these results rule out rigid template wording and single-token answer scoring as
explanations for the behavioural window, while keeping the stronger copying-to-suppression
claims scoped to the interventions tested in the main analysis.

\section{The Second-Mention State and Input Controls}

\label{app:s2}

\subsection{Position controls}
\begin{figure*}[htbp]
\centering
\includegraphics[width=\textwidth]{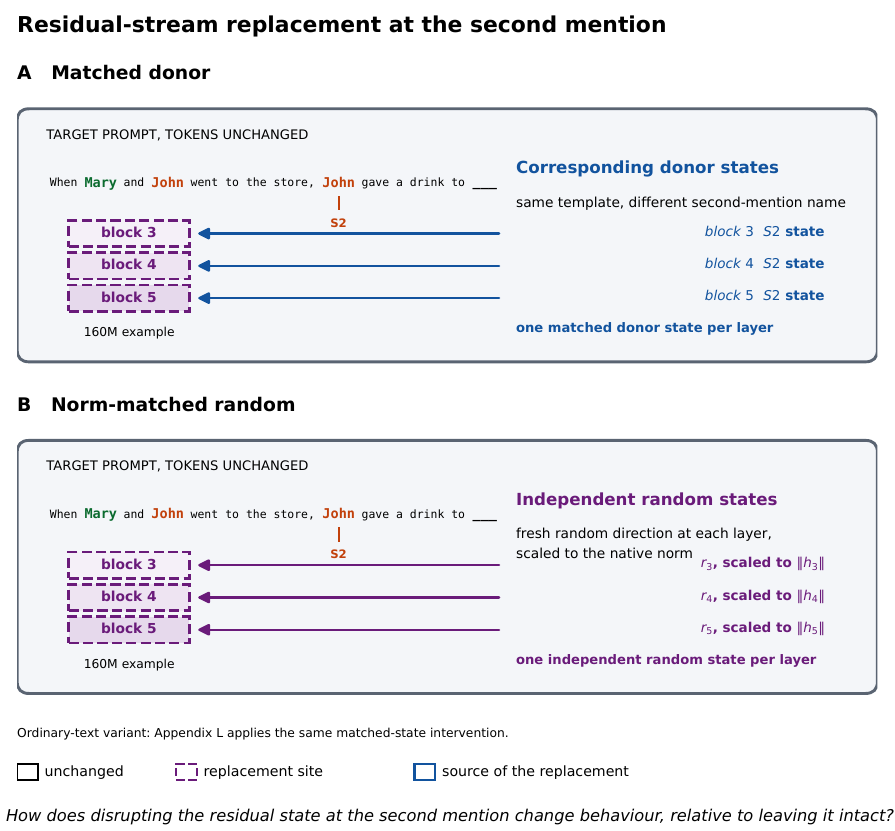}
\caption{Residual-stream replacement at the second mention, in two variants. Neither edits the evaluated prompt. Only the activation carried at that position is replaced, after each block in an early layer band. Panel A takes the replacement from a matched control prompt that differs only in the name at the second mention, so the inserted state carries another identity. Panel B instead samples a fresh independent direction for every prompt and every layer, scaling each to the norm of the state it replaces, so no name from any prompt enters. Comparing the two separates disruption of the state from content imported with it.}
\label{fig:atlasCD}
\end{figure*}

\begin{figure*}[htbp]
\centering
\includegraphics[width=\textwidth]{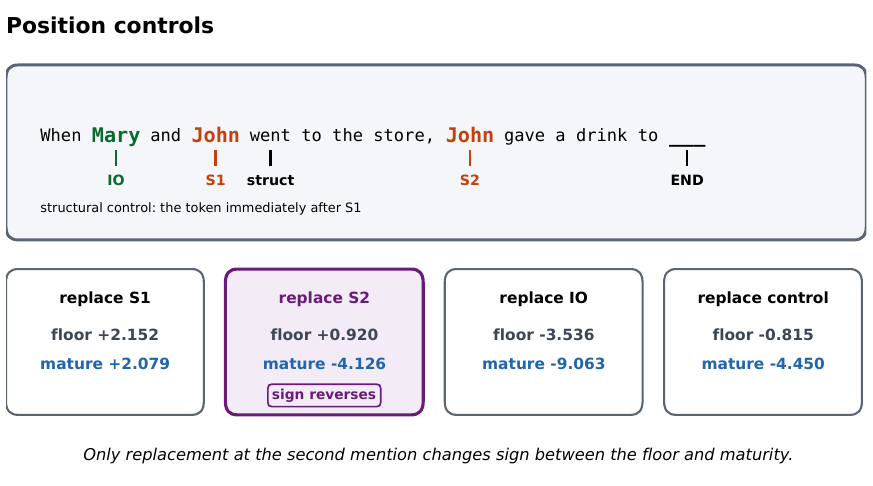}
\caption{Position controls. The norm-matched random replacement of Figure~\ref{fig:atlasCD} is applied separately at four sites in the same sentence. Values are the change in logit difference at the 160M floor and at maturity. The structural control is defined positionally, as the token immediately following the first subject mention, so the word marked in this example sentence varies across sentence templates. Replacement at the first mention increases the logit difference at both checkpoints, whereas replacement at the indirect object and at the structural control decreases it at both. Only replacement at the second mention changes sign. This is what makes the second-mention state specific rather than one of several positions carrying a causal effect.}
\label{fig:atlasE}
\end{figure*}

The replaced activations are the residual-stream outputs of an early band of transformer blocks. The band covers layers 3 to 5 of 12 at 160M, layers 6 to 10 of 24 at 410M, layers 4 to 8 of 16 at 1B, and layers 8 to 14 at 2.8B and 6.9B, and layers 9 to 16 at 12B. At 6.9B and 12B, forced-choice accuracy falls below chance at step 2000, to $0.323$ and $0.420$. Matched second-mention state replacement changes the logit difference by $+0.837$ and $+0.434$ at those floors, and by $-4.118$ and $-3.955$ at step 143000. A separate replacement vector is drawn for every prompt and every layer in the band. Figure~\ref{fig:atlasCD} sets the two variants side by side on the canonical prompt. Figure~\ref{fig:atlasE} marks the four tested sites on one sentence.

Table~\ref{tab:s2battery} reports the controls. Replacement at the second subject mention improves the logit difference by $0.953$ at the floor and harms it by $4.123$ at maturity. Replacing the same activations with a Gaussian vector rescaled to the magnitude of what it replaces, drawn independently per prompt, gives $0.920$ and $-4.126$. The two agree to within a few hundredths at both checkpoints.

That agreement determines what the experiment establishes. The random direction is sampled independently while its norm is taken from the state it replaces, so it imports no content from another prompt. Its reproducing the matched effect therefore shows the result depends on disrupting the activation rather than on supplying different name content. The remaining positions show no sign reversal.

\begin{table}[htbp]
\centering\small
\begin{tabular}{@{}lcc@{}}
\toprule
Position and replacement & Floor & Mature \\
\midrule
Second mention, matched & $+0.953$ & $-4.123$ \\
Second mention, random & $+0.920$ & $-4.126$ \\
First mention, random & $+2.152$ & $+2.079$ \\
Indirect object, random & $-3.536$ & $-9.063$ \\
Structural position, random & $-0.815$ & $-4.450$ \\
\bottomrule
\end{tabular}
\caption{Change in logit difference from state replacement at each position, at the 160M floor and mature checkpoints.}
\label{tab:s2battery}
\end{table}

\subsection{Sensitivity to the layer band}
\label{app:bands}

The main 160M intervention uses layers 3 to 5. We test whether the early effect depends strongly on this choice using the separately trained 160M model at step 2000. This checkpoint precedes that model's floor at step 2850, so we use this experiment only as a sensitivity check on the layer choice rather than as a separate test of the floor-to-mature reversal.

We divide the model's 12 layers into four non-overlapping three-layer bands and repeat the same matched $S2$ activation-patching intervention in each band. The change in logit difference is $+0.447$ for layers 0--2, $+0.388$ for layers 3--5, $+0.443$ for layers 6--8, and $+0.133$ for layers 9--11. The corresponding $95\%$ intervals are $[0.378, 0.520]$, $[0.322, 0.458]$, $[0.395, 0.496]$, and $[0.113, 0.153]$. Each band is evaluated on 300 prompts.

The intervention has a positive effect across all four parts of the network rather than only in the layer band used for the main analysis, although the effect is weaker in the final three layers. This sweep uses its own evaluation set of ten templates with thirty prompts each, so its baseline logit difference differs from that of the main analysis.

\subsection{Removing the repetition at the input}

\begin{figure}[htbp]
\centering
\includegraphics[width=\columnwidth]{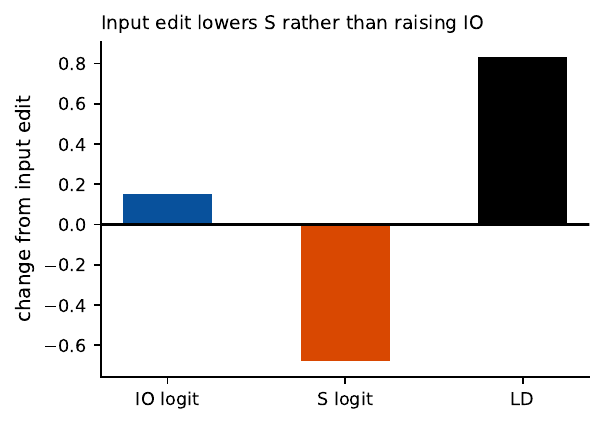}
\caption{Decomposition of the input edit at the 160M floor.}
\label{fig:appinput}
\end{figure}

A complementary experiment changes the tokens. For each prompt the second occurrence of the subject is replaced by a third single-token name differing from both the subject and the indirect object, which preserves sequence length and every token position while removing the repeated-name relation. On a fixed evaluation set of 800 prompts at the 160M floor this improves the logit difference by $0.833$, with a crossed template and name-pair interval of $[0.681, 1.023]$, and moves forced-choice accuracy from $0.309$ to $0.504$. The change is driven mainly by lowering the repeated subject, since the subject logit falls by $0.681$ while the object logit rises by $0.152$, as Figure~\ref{fig:appinput} shows. Because the edit changes repetition status and token identity together, it gives repetition-sensitive evidence without isolating repetition from lexical identity. A control substituting a different third name gives a very similar improvement, and a control editing a nearby filler token gives a much smaller effect.

\section{Copying-Head Interventions and Controls}
\label{app:copying}

This appendix gives the full protocol behind the copying results in
Section~\ref{sec:copying}. We first explain how the head set is selected and how the data
used for selection, replacement construction, and evaluation are kept separate. We then
give the complete control-mean intervention, the uncertainty analyses, the source-identity
experiment, independent functional checks on the heads, and a final comparison that
measures copying and suppression using the same intervention.

\subsection{Head selection and data separation}
\label{app:design}
\begin{figure*}[htbp]
\centering
\includegraphics[width=\textwidth]{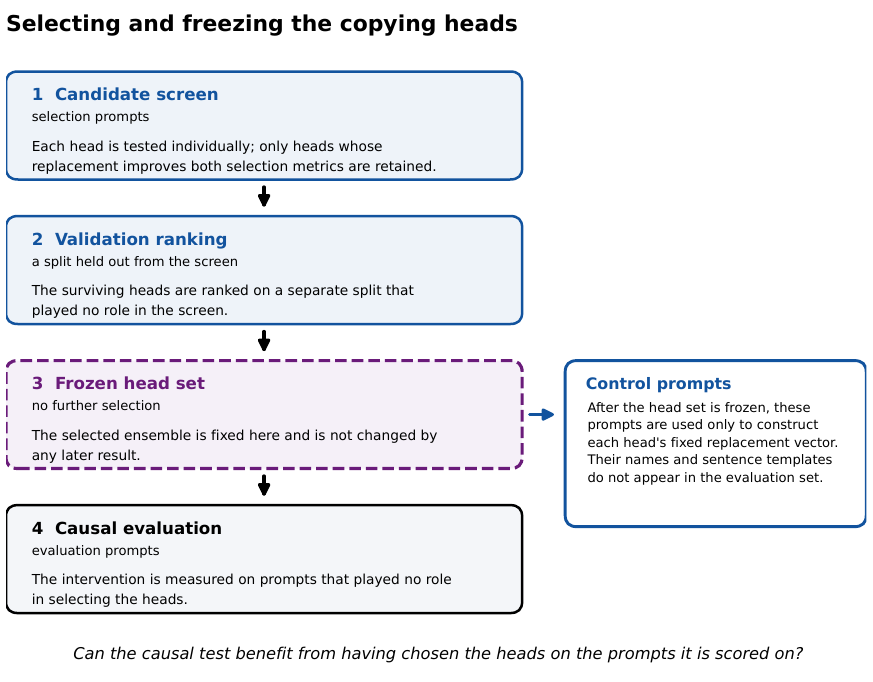}
\caption{Selecting and freezing the copying heads. Four prompt resources play four distinct roles: a selection set for the candidate screen, a validation split held out from that screen, a control-mean set used only after freezing to build the replacement vectors, and an evaluation set. Candidate heads are kept when replacing each on its own improves both the three-name margin and the correct-name probability. The survivors are then ranked on a validation split that took no part in that screen, and the resulting set is frozen and never reselected. Table~\ref{tab:headsets} lists the frozen set for every model. Only afterwards is the intervention effect measured, on evaluation prompts whose names and sentence templates never influenced selection.}
\label{fig:atlasG}
\end{figure*}

Selection proceeds in three stages on disjoint prompts. We first screen candidates by replacing each head individually on a selection set. A head qualifies only if replacing it improves both the three-name margin and the correct name's probability. We use only the sign of each change at this stage rather than its magnitude, and we do not rank heads by how strongly they write to the output logits. This makes the initial screen intervention-based rather than attribution-based and avoids choosing a magnitude threshold. We then rank the surviving heads by their replacement effect on a separate validation split that took no part in that screen. The resulting set is frozen before any reported effect is
measured, and no later experiment changes which heads are intervened on. Figure~\ref{fig:atlasG}
summarises the four stages and the prompt sets each one uses.

After the head set has been fixed, the remaining names and sentence templates are divided
again into two non-overlapping groups. The control-mean group is used only to construct
the replacement activation for each selected head. The evaluation group is used only to
measure the resulting causal effect. No names or sentence templates are shared between
these two groups.

The control-mean prompts are non-repeated variants in which the two relevant name
positions contain different names. Both name orders are included so that the resulting
average is not tied to one name occupying one fixed role. The held-out evaluation prompts
contain the ordinary repeated-name structure.

\begin{figure*}[htbp]
\centering
\includegraphics[width=\textwidth]{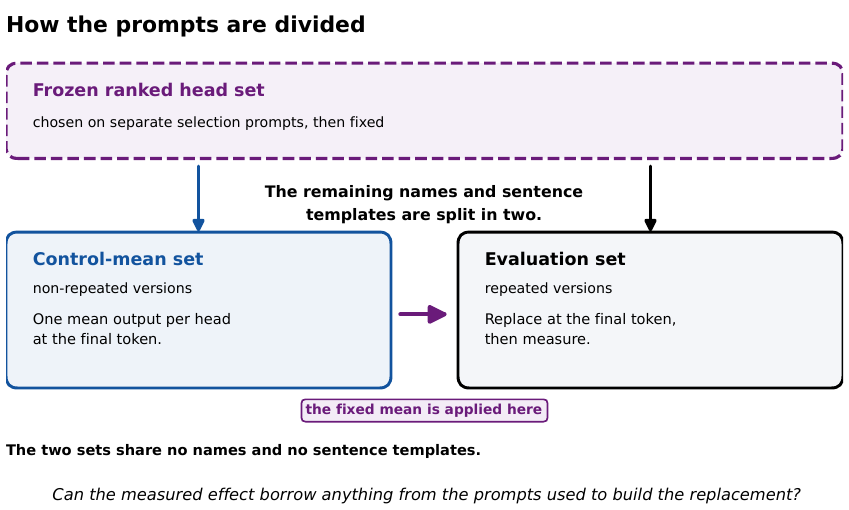}
\caption{How the prompts are divided. After the head set is frozen, the remaining names and sentence templates are split into a control-mean set of non-repeated prompts, used only to build one fixed mean output per head, and an evaluation set of repeated-name prompts on which the effect is measured. The two share no names and no sentence templates, so the measured effect cannot borrow from the prompts that produced the replacement.}
\label{fig:appdesign}
\end{figure*}

The protocol therefore separates three roles. One set of prompts is used to select and
freeze the head ensemble; a separate set of non-repeated prompts is used to construct the
fixed replacement vectors; and held-out repeated-name prompts are used only to measure the
final causal effect. Figure~\ref{fig:appdesign} summarises this separation. Keeping the
names and sentence templates disjoint prevents either head selection or replacement
construction from benefiting from the examples on which the effect is finally measured.

This separation matters because the causal test should not benefit from selecting heads, constructing their replacement, and evaluating them on the same names or sentence templates.

\begin{table*}[htbp]
\centering\footnotesize
\setlength{\tabcolsep}{3pt}
\begin{tabular}{@{}l c c c >{\raggedright\arraybackslash}p{8.2cm}@{}}
\toprule
Setting & Control mean & Evaluation & $n$ & Frozen head set \\
\midrule
retrained 160M & 64 & 192 & 12 & \footnotesize L2H11, L5H0, L5H7, L6H8, L7H1, L8H2, L9H1, L9H10, L10H4, L10H10, L11H10, L11H11 \\
official 160M & 64 & 192 & 21 & \footnotesize L2H10, L3H10, L4H0, L4H6, L4H8, L4H11, L5H5, L5H9, L6H0, L6H6, L7H6, L7H10, L8H1, L8H2, L9H3, L9H6, L9H7, L9H8, L10H1, L10H7, L11H11 \\
official 410M & 64 & 192 & 17 & \footnotesize L11H8, L19H10, L17H6, L13H5, L16H15, L12H4, L17H13, L10H14, L16H2, L14H7, L14H0, L20H4, L10H0, L10H9, L16H7, L11H14, L11H1 \\
official 1B & 64 & 192 & 22 & \footnotesize L5H6, L6H7, L9H6, L6H4, L7H1, L4H0, L5H7, L8H5, L11H2, L15H1, L11H1, L14H2, L5H2, L14H5, L15H0, L12H2, L4H1, L13H3, L3H1, L3H4, L13H4, L4H5 \\
\bottomrule
\end{tabular}
\caption{Sizes in items and the frozen head sets. An item is a name pair evaluated in both orders. The control-mean half and the evaluation half share no names and no templates. Heads are written as layer and index, and the sets are screened, then ranked on a separate validation split, then frozen before the mean is estimated or the effect is measured.}
\label{tab:headsets}
\end{table*}

\subsection{Control-mean replacement}
For each frozen attention head we compute one fixed replacement vector by averaging that
head's output at END over the non-repeated prompts in the control-mean group. The average
spans multiple names, sentence templates, and both name orders, so it does not correspond
to the activation produced by any particular donor prompt or alternative answer.

On each held-out repeated-name prompt we replace the selected heads' prompt-specific
outputs at END with these fixed control means. The rest of the model then proceeds
normally.

\begin{figure*}[htbp]
\centering
\includegraphics[width=\textwidth]{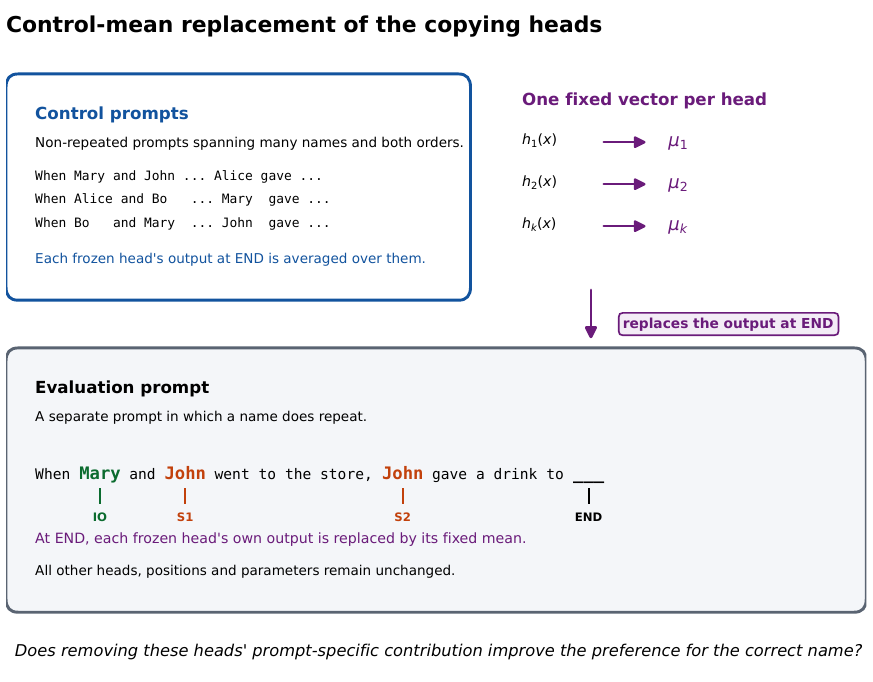}
\caption{Control-mean replacement of the copying heads. The replacement is one fixed vector per head, obtained by averaging that head's output at the final token over many non-repeated control prompts spanning several names and both name orders. It is therefore not an activation borrowed from a single donor prompt, and it carries no particular competing name. On a separate evaluation prompt in which a name does repeat, each frozen head's own output is replaced by its fixed mean while every other head, position and parameter is left alone.}
\label{fig:atlasH}
\end{figure*}

The intervention is deliberately different from replacing an activation with one taken
from another prompt of the same task. A single donor prompt can carry the identity of a
particular alternative name and thereby improve the prediction simply by introducing that
name. Averaging across many non-repeated controls avoids assigning the replacement to one
specific donor identity. We do not need the stronger assumption that every possible
name-specific feature cancels exactly in the average. The relevant property is narrower:
the replacement is not the activation from a particular prompt carrying a particular
competing name. Figure~\ref{fig:atlasH} follows one head from the control prompts to its fixed mean and then to the evaluation prompt.

We evaluate the intervention with two outcome measures. The first is the change in the
$IO$ versus $S$ logit difference, $\Delta\ld = \ld_{\text{replaced}} - \ld_{\text{native}}$,
where a positive value means that replacing the heads' native outputs improves the
model's preference for the correct name relative to the repeated name. The second is the
three-name margin of equation~\ref{eq:three-name-margin}, where $ALT$ is an unrelated third name. The stricter
measure tests whether an apparent improvement over $S$ remains when another plausible
named competitor is included.

\begin{figure*}[htbp]
\centering
\includegraphics[width=\textwidth]{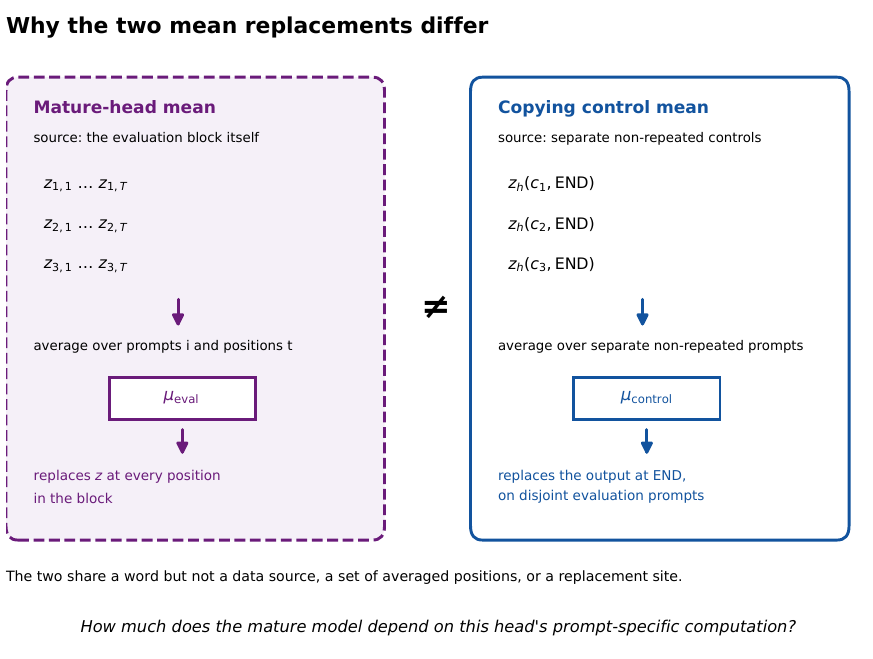}
\caption{Mature-head mean replacement, and how it differs from the control mean of Appendix~\ref{app:copying}. Here the selected head's activation is replaced by its own mean over the prompts and token positions of each sentence-template block being evaluated, so the mean comes from the evaluation data itself. In the copying experiment the mean is instead built from a separate set of non-repeated control prompts. The two operations share a word but not a data source, and they answer different questions. Because this head was also chosen by its replacement effect, the size of that effect is selection-conditioned.}
\label{fig:atlasK}
\end{figure*}

This control-mean replacement should not be confused with the mature-head mean replacement
used in Section~\ref{sec:suppression}. Here the mean is constructed from a separate set of
non-repeated control prompts. There, the selected mature suppressor's activation is
replaced by its mean over the prompts and token positions of each sentence-template block. They are different interventions answering
different causal questions. Figure~\ref{fig:atlasK} places this mean beside the copying control mean, which is built from different data.

\subsection{Replacing the selected heads improves the floor prediction}

\begin{table}[t]
\centering\footnotesize
\setlength{\tabcolsep}{3pt}
\begin{tabular}{@{}lcc@{}}
\toprule
Setting at floor & $\Delta\ld$ & $\Delta\margin$ \\
\midrule
retrained 160M & $+0.180$ & $+0.115$ \\
 & \scriptsize$[0.004, 0.362]$ & \scriptsize$[-0.072, 0.292]$ \\[2pt]
official 160M & $+0.804$ & $+0.501$ \\
 & \scriptsize$[0.529, 1.132]$ & \scriptsize$[0.209, 0.846]$ \\[2pt]
official 410M & $+0.285$ & $+0.226$ \\
 & \scriptsize$[0.179, 0.388]$ & \scriptsize$[0.103, 0.345]$ \\[2pt]
official 1B & $+0.381$ & $+0.139$ \\
 & \scriptsize$[0.165, 0.640]$ & \scriptsize$[-0.064, 0.404]$ \\
\bottomrule
\end{tabular}
\caption{Effect of the control-mean output replacement, in which each frozen head's prompt-specific output at the final token is replaced by its average output over non-repeated control prompts, with nested resampling intervals. Positive means the preference for the correct name improved.}
\label{tab:m1}
\end{table}
Control-mean replacement improves the $IO$ versus $S$ logit difference at all four
evaluated floors. The effects are $+0.180$ for the separately trained 160M model, $+0.804$
for released 160M, $+0.285$ for 410M, and $+0.381$ for 1B. Under template-to-item nested
resampling all four intervals remain above zero. Table~\ref{tab:m1} gives both measures
with their intervals.

The stricter three-name margin has a positive point estimate in all four settings. Its
template-aware interval lies above zero for released 160M and 410M, and crosses zero for
the separately trained 160M model and for 1B. The uniform conclusion is therefore the
two-name result: replacing the selected heads' prompt-specific outputs improves the
model's preference for the correct name relative to the repeated name.

We additionally report item-clustered intervals, which treat paired contrast rows from the
same item as one resampling unit. These are generally narrower than the template-to-item
intervals because the latter additionally account for the small number of shared
sentence-template groups. Both analyses and their precise resampling definitions are given
in Appendix~\ref{app:protocols}. Figure~\ref{fig:appm1unc} compares the two procedures, and Table~\ref{tab:m1full} gives every cell.

These interventions identify a causally harmful contribution from the selected ensemble at the floor. Appendix~\ref{app:copying-source-identity} next tests whether that contribution follows the identity supplied at the heads’ source position.

\paragraph{Effect later in training.} We apply the same frozen head set and the same
replacement procedure at later checkpoints. At recovery and at maturity the strong
positive effect seen at the floor becomes much smaller and is no longer consistently
positive. The harmful contribution is therefore not a fixed property of these heads
throughout training. The importance of a head depends on the surrounding computation in
which it participates, which is one reason a head's role at a single checkpoint should not
automatically be treated as its role throughout training.

\FloatBarrier
\subsection{Difference between the two-name and three-name measures}
The two measures answer related but different questions. The two-name measure asks whether
replacing the selected heads improves the correct name relative specifically to the
repeated name. The three-name margin additionally requires the correct name to improve
relative to an unrelated third candidate. An intervention can therefore strongly improve
$IO$ minus $S$ while producing a smaller change in the three-name margin, if some of the
changed logit mass is redistributed toward the third name.

For this reason the two quantities should not be read as interchangeable estimates of one
underlying effect. The two-name result is the direct test of the repeated-name preference.
The three-name result is a stricter robustness check against an improvement that is
specific only to that comparison.

The same distinction matters when comparing this experiment with the source-identity
intervention below. The two experiments use the same outcome metric while changing
different parts of the model, and therefore estimate different causal effects. Their
magnitudes should not be compared as though one were a stronger version of the other.

\begin{figure}[htbp]
\centering
\includegraphics[width=\columnwidth]{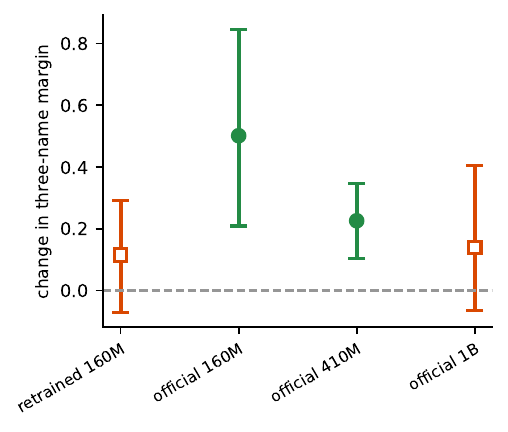}
\caption{Change in the three-name margin of equation~\ref{eq:three-name-margin} from the control-mean replacement. Filled circles indicate that the template-to-item nested interval lies above zero and open squares that it crosses zero. Under item-clustered resampling all four intervals lie above zero.}
\label{fig:appthreename}
\end{figure}

\begin{figure*}[t]
\centering
\includegraphics[width=0.86\textwidth]{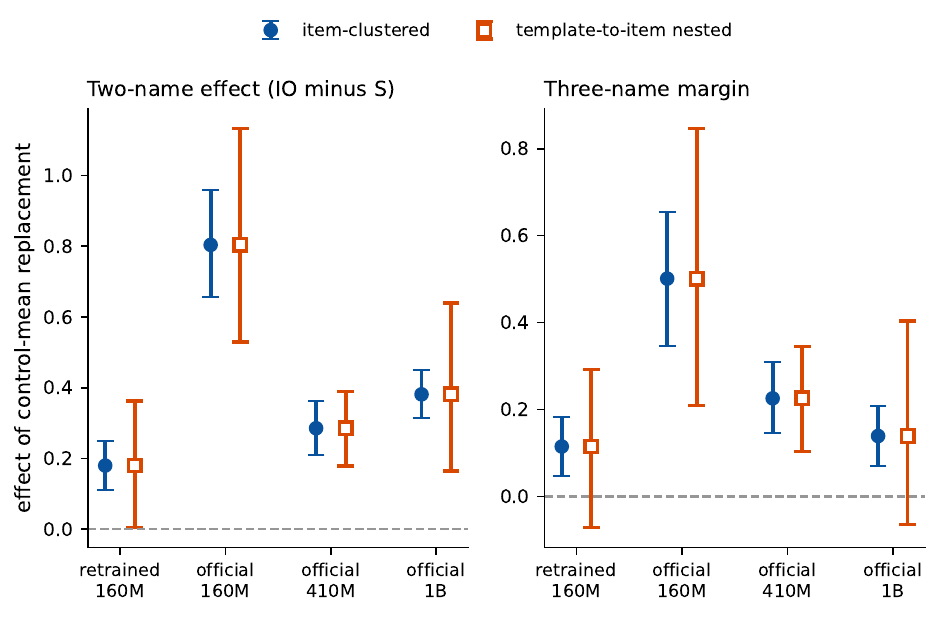}
\caption{Both measures under both resampling procedures. Item-clustered intervals are narrower because the nested procedure additionally resamples a small number of template groups.}
\label{fig:appm1unc}
\end{figure*}

\subsection{The contribution follows the identity supplied at the first mention}
\label{app:copying-source-identity}
The control-mean experiment shows that replacing the selected heads' outputs improves
behaviour, but it does not by itself establish what information those outputs carry. We
therefore perform a more targeted intervention that asks whether the heads' downstream
contribution follows the identity of the name they read.

For every target prompt we construct a donor prompt with the same name at $S2$ but a
different name at $S1$. The two contrasts can be written schematically as $XX \leftarrow YX$
and $YY \leftarrow XY$. The repeated identity at $S2$ is therefore held fixed while the
identity available at the earlier source position changes.

We then replace only the selected heads' value vectors at $S1$ with the corresponding
value vectors from the donor prompt. The target prompt itself remains unchanged, its $S2$
token remains unchanged, the rest of the model remains native, and downstream computation
proceeds from the modified values.

\begin{figure*}[htbp]
\centering
\includegraphics[width=\textwidth]{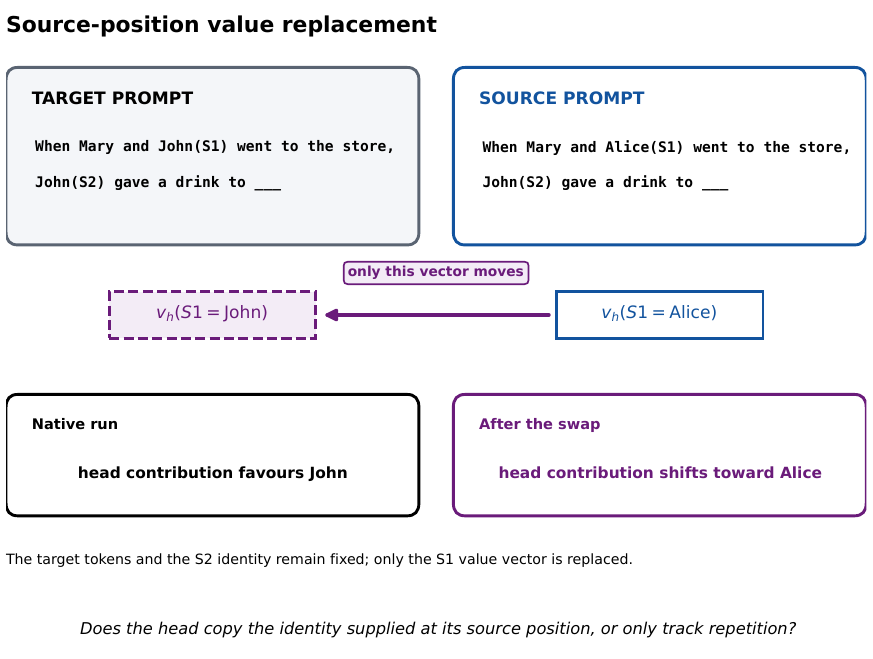}
\caption{Source-position value replacement. The evaluated prompt and the identity at the second mention remain unchanged. For each frozen head, only the value vector read at the first mention is replaced by the corresponding value from a prompt carrying a different name there. If the heads merely wrote toward whichever name repeats, their downstream contribution would be unaffected; if it instead follows the substituted identity, the heads are carrying the name represented at their source. This distinction is what the term copying head refers to.}
\label{fig:atlasI}
\end{figure*}

This asks a more specific question than the control-mean replacement. If the heads merely
correlate with a high repeated-name logit, changing the identity supplied at $S1$ need not
make their downstream contribution follow the substituted name. If they carry name
identity from $S1$, however, substituting the value there should redirect which name they
support. Figure~\ref{fig:atlasI} makes the intervention concrete by placing the target and source prompts side by side and marking the single value vector that is transferred.

Instead, the contribution follows the substituted identity. At every evaluated floor the original repeated name's logit falls
while the substituted name's logit rises, in the expected opposing directions, and the
three-name margin improves. Table~\ref{tab:valuecopy} gives the component logit changes
and Table~\ref{tab:vs1} the three-name margin under both uncertainty procedures.

\begin{table*}[t]
\centering\small
\setlength{\tabcolsep}{4pt}
\begin{tabular}{@{}lcccc@{}}
\toprule
Setting & Heads & Repeated & Substituted & Three-name margin \\
\midrule
retrained 160M & 12 & $-0.478$ & $+0.550$ & $+0.412$ \\
official 160M & 21 & $-2.025$ & $+2.257$ & $+0.972$ \\
official 410M & 17 & $-0.732$ & $+0.754$ & $+0.596$ \\
official 1B & 22 & $-0.364$ & $+0.417$ & $+0.212$ \\
\bottomrule
\end{tabular}
\caption{Component logit changes from replacing only the frozen ensemble's value vectors at the first subject mention, holding the second mention fixed. All four settings use the same source-position value-replacement protocol. The repeated-name logit falls and the substituted name's logit rises by a comparable amount at every setting, and the three-name margin improves at every setting. Table~\ref{tab:vs1} gives that change under both uncertainty procedures.}
\label{tab:valuecopy}
\end{table*}

\begin{table}[htbp]
\centering\small
\setlength{\tabcolsep}{4pt}
\begin{tabular}{@{}lccc@{}}
\toprule
Setting & Change & Item-clustered & Template-nested \\
\midrule
retrained 160M & $+0.412$ & \footnotesize$[0.370,0.455]$ & \footnotesize$[0.313,0.518]$ \\
official 160M & $+0.972$ & \footnotesize$[0.898,1.047]$ & \footnotesize$[0.806,1.137]$ \\
official 410M & $+0.596$ & \footnotesize$[0.542,0.650]$ & \footnotesize$[0.436,0.742]$ \\
official 1B & $+0.212$ & \footnotesize$[0.179,0.248]$ & \footnotesize$[0.128,0.315]$ \\
\bottomrule
\end{tabular}
\caption{Change in the three-name margin from replacing the frozen ensemble's value vectors at the first subject mention, under both uncertainty procedures. Every interval lies above zero. This experiment carries six to eight sentence templates, compared with three or four for the control-mean experiment, which is why its nested intervals widen less relative to the item-clustered ones.}
\label{tab:vs1}
\end{table}

The uncertainty calculation first averages the two same-$S2$ contrast rows belonging to
each item and then resamples those item-level values. The template-aware analysis
additionally resamples the sentence-template groups before resampling items within them.

The key result is not simply that replacing these heads helps. Their downstream
contribution changes identity when the identity supplied at their source position changes.
This is the evidence that licenses describing the ensemble as performing repeated-name
copying rather than merely correlating with the repeated-name logit.

\subsection{Independent functional evidence from subject-writing heads}
The source-identity intervention above is the direct causal evidence for copying. We also
perform a separate functional analysis that does not use that intervention to choose the
heads.

At each model size we identify the attention heads that most strongly write the prediction
toward the repeated subject at the floor, ranked by their signed direct contribution to
the $IO$ minus $S$ logit difference. We then inspect two properties expected of a copying
head: whether the head attends from END to positions containing the repeated subject, and
whether its value-output transformation maps the attended subject representation toward
that subject's output direction. These are descriptive diagnostics applied after the
subject-writing heads have been identified.

For released Pythia 160M the five leading subject writers are L5H9, L10H7, L5H5, L4H6, and
L9H7. Four of the five satisfy the resulting copy-like diagnostic. L5H9, for example, has
a subject-copying output score of $0.674$ and places substantial attention on the
repeated-name positions. At 410M all five leading subject writers satisfy the criterion,
and at 1B all five do as well.

These diagnostics provide corroborating evidence. A head can have an
output direction that decodes as copying without that decoded direction being the causal
quantity responsible for behaviour. Appendix~\ref{app:alternatives} shows that directly
removing a linearly decoded duplication direction makes floor behaviour worse rather than
better. The causal copying claim therefore rests on the output replacement and the
source-identity intervention, while this analysis provides independent evidence that heads
identified because they write toward the repeated subject have the functional properties
expected of name-copying heads.

\subsection{Copying and suppression measured with the same intervention}
The main copying and suppression analyses use different interventions because they answer
different causal questions. As an additional comparison we therefore measure frozen
copying and suppressing ensembles using the same intervention and the same outcome metric.

Two head sets are used. The first is the frozen floor copying ensemble described above. The
second is a separate mature suppressor ensemble, which is not the single mature suppressing
head analysed in Section~\ref{sec:suppression}. It is obtained from an independent
all-head scan at the mature checkpoint in which candidate heads are ranked by the effect of
replacing their full output at END using a donor that preserves the $S2$ identity while
changing the earlier name. Candidate suppressors must then reproduce the suppressive sign
on a separate validation split, must act by reducing support for the repeated name under
the native model, and must have the same sign in both identity-swap directions. The
selection and validation criteria are therefore applied before the developmental
comparison. This yields 15 validated mature suppressor heads for the separately trained
160M model and 14 for the released 160M model. Neither ensemble is reselected as training
proceeds.

For each target prompt the donor keeps the identity at $S2$ fixed and changes the identity
at $S1$. We then replace the selected ensemble's full attention-head outputs at END with
the outputs produced by that donor, applying the same operation separately to the copying
ensemble and to the mature suppressor ensemble. The outcome is the change in the three-name
margin. Under this sign convention a positive change means replacement improves behaviour,
so the native ensemble was harmful, and a negative change means replacement harms
behaviour, so the native ensemble was helping.

\begin{figure*}[t]
\centering
\includegraphics[width=0.94\textwidth]{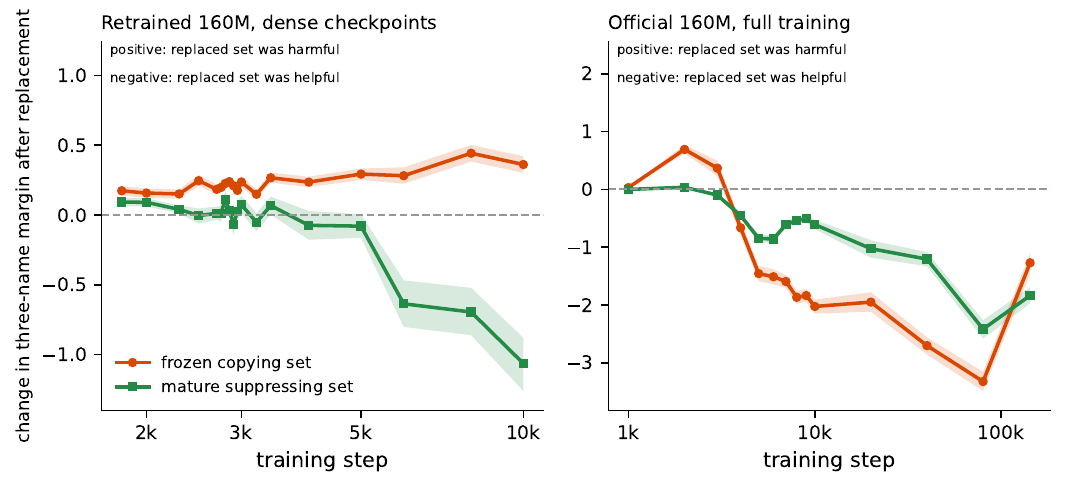}
\caption{The two computations measured on one scale. Each curve is the change in the three-name margin when the same intervention is applied to one of the two frozen head sets, with shaded 95 percent intervals. The left panel uses the retrained 160M model with dense checkpoints through the transition and the right panel uses the released 160M model over full training. A positive value means replacing that set improved behaviour, so the native set was working against the correct answer. A negative value means replacing it damaged behaviour, so the native set was helping.}
\label{fig:trajectory}
\end{figure*}

In the densely checkpointed separately trained 160M model, the copying ensemble already
has a reliably harmful contribution at the earliest evaluated checkpoint. The mature
suppressor ensemble becomes reliably helpful only later. The trajectory identifies
the first positive-interval copying effect at step 1800 and the first reliably negative
suppressor effect at step 6000. These are checkpoint-resolution markers rather than exact
circuit birth times.

The released 160M model shows the same qualitative ordering, although the later trajectory
of the frozen floor copying set differs. In that model the originally harmful set
eventually changes causal sign and becomes useful rather than remaining a fixed harmful
pathway. This difference is informative rather than contradictory, since component roles
can themselves change during training.

The common result is the developmental ordering. Figure~\ref{fig:trajectory} traces both ensembles. A repeated-name-writing contribution is
already harmful during the early failure, while the ensemble selected for mature
suppression becomes strongly helpful later.

\subsection{Evidence for harmful repeated-name copying}
The experiments in this appendix progressively narrow the interpretation. Control-mean
replacement shows that the selected heads' prompt-specific outputs causally worsen the
floor preference for the correct name relative to the repeated name. The source-identity
intervention then shows that their contribution follows the identity supplied at the
position from which they read. Independently selected subject-writing heads also
predominantly exhibit the attention and output properties expected of copying heads.
Finally, measuring frozen copying and suppressing ensembles under a common intervention
shows that their causal importance develops at different times.

Together these results support a causally harmful repeated-name copying ensemble at the
floor. The selected ensemble need not be unique or complete, and individual heads within it may differ in their precise roles. The later suppression analysis is therefore kept conceptually separate:
Section~\ref{sec:suppression} asks how a mature suppressing computation develops and
whether giving that computation to the early model can improve its behaviour.

\begin{table*}[t]
\centering\small
\setlength{\tabcolsep}{4pt}
\begin{tabular}{@{}>{\raggedright\arraybackslash}p{2.1cm}>{\raggedright\arraybackslash}p{1.5cm}ccc>{\raggedright\arraybackslash}p{1.7cm}@{}}
\toprule
Setting & Measure & Point estimate & Item-clustered interval & Nested interval & Resolves under nested \\
\midrule
retrained 160M & two-name & $+0.180$ & $[0.110,0.248]$ & $[0.004,0.362]$ & yes \\
 & three-name & $+0.115$ & $[0.047,0.184]$ & $[-0.072,0.292]$ & no \\
official 160M & two-name & $+0.804$ & $[0.657,0.958]$ & $[0.529,1.132]$ & yes \\
 & three-name & $+0.501$ & $[0.346,0.655]$ & $[0.209,0.846]$ & yes \\
official 410M & two-name & $+0.285$ & $[0.210,0.362]$ & $[0.179,0.388]$ & yes \\
 & three-name & $+0.226$ & $[0.146,0.309]$ & $[0.103,0.345]$ & yes \\
official 1B & two-name & $+0.381$ & $[0.315,0.450]$ & $[0.165,0.640]$ & yes \\
 & three-name & $+0.139$ & $[0.069,0.208]$ & $[-0.064,0.404]$ & no \\
\bottomrule
\end{tabular}
\caption{Both measures under both resampling procedures, as 95 percent intervals from ten thousand draws.}
\label{tab:m1full}
\end{table*}

\section{Specificity of the Selected Copying Heads}
\label{app:specificity}
The copying experiment establishes that replacing the selected heads improves behaviour at
the floor. A separate question is whether this result is specific to the selected
ensemble, or whether almost any set of heads with similar observable properties would
produce the same effect.

We test this using feature-matched control sets. For each selected head we construct
matched alternatives from other heads in the same layer. Matching uses six quantities
measured on a resource-disjoint selection split: the attention from END to the first
subject mention, to the second, and to the indirect object; the norm of the head's output
at END; its direct projection toward the repeated name; and the selectivity of its value
representation at the first mention for the identity supplied there. We standardise these
six features within each layer and rank candidate controls by Euclidean distance from the
selected head in that standardised space. The selected heads themselves are excluded. We
then construct 200 control ensembles with the same number of heads as the selected
ensemble and evaluate every ensemble using the same intervention and the same held-out
evaluation prompts.

These controls are more demanding than random-head controls. They ask whether the causal
result can be explained simply by choosing heads that attend to names and write strongly
at the final position, rather than by the particular heads our selection procedure
identifies. The specificity experiment also uses names and sentence templates separated
from those used to identify the ensemble.

The selected-set effects here should not be expected to equal those in
Appendix~\ref{app:copying} exactly. This experiment uses its own held-out prompts so that
the selected set and all 200 matched controls are compared on identical examples. The two
appendices therefore estimate the same kind of intervention effect on different evaluation
sets.

At the released 160M floor the selected ensemble is clearly separated from the matched
controls. Its two-name effect is $+1.092$, while the 200 matched control sets range from
$-0.023$ to $+0.056$. On the three-name margin the selected effect is $+0.410$, against a
control range of $-0.023$ to $+0.054$. No matched set approaches the selected ensemble on
either measure, giving an empirical upper-tail probability of $0.005$ for both.

The separation is narrower in the separately trained 160M model. The selected set gives
$+0.319$ on the two-name measure while the matched controls extend as high as $+0.316$, an
empirical upper-tail probability of $0.005$. On the three-name margin the selected effect
is $+0.240$ while the controls reach $+0.250$, so at least one matched set exceeds the
selected ensemble and the empirical upper-tail probability is $0.015$, placing the selected
set at the 99th percentile rather than beyond every control.

The two settings therefore support different strengths of conclusion. At released 160M the
selected ensemble is strongly separated from heads matched on these observable properties.
At the separately trained model it is unusual but not cleanly separated from the strongest
matched alternatives, particularly on the three-name measure. This agrees with the wider
uncertainty seen for that setting in Appendix~\ref{app:copying}. Figure~\ref{fig:appnull} plots the selected set against the control distribution.

These controls show that the selected ensemble is specific relative to the tested feature-matched control family. The separation is strong in the released 160M model and narrower in the separately trained model, particularly for the three-name margin. The experiment establishes that the result is not produced by an arbitrary collection of heads with similar observable properties.

\begin{figure*}[t]
\centering
\includegraphics[width=0.92\textwidth]{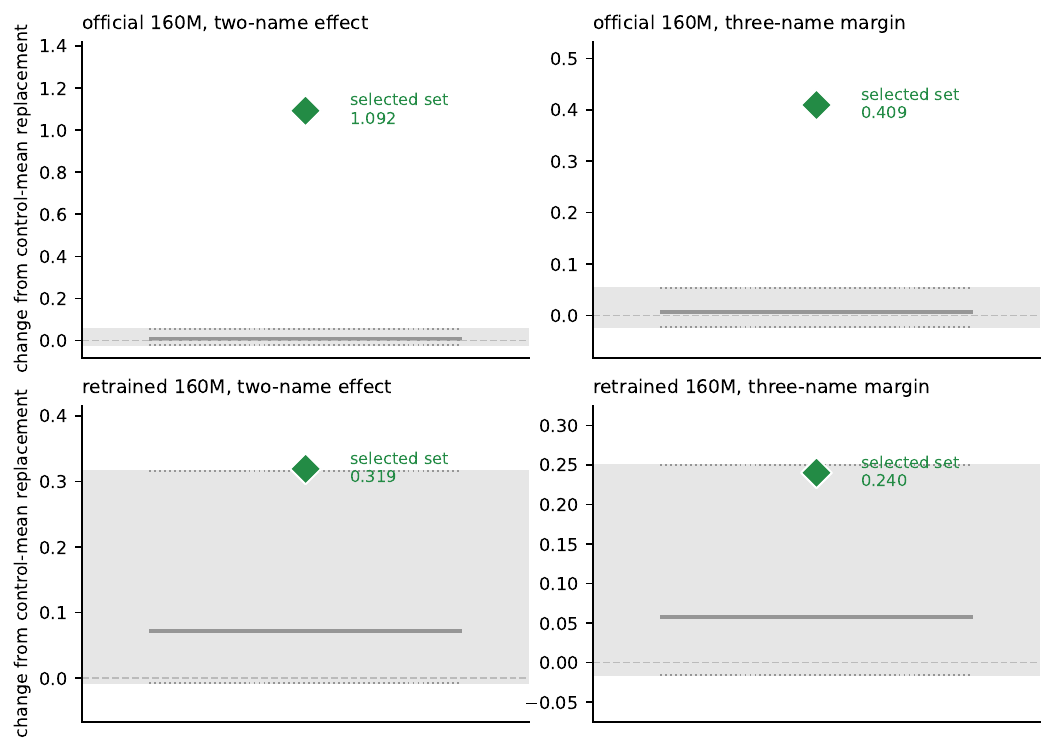}
\caption{The selected head set against two hundred feature-matched control sets, at the official 160M floor in the upper row and the retrained 160M floor in the lower row. The shaded band spans the full range of the two hundred control-set effects and the solid line marks their mean, with the diamond marking the selected set. The control ranges are $-0.023$ to $+0.056$ and $-0.023$ to $+0.054$ at the official model, and $-0.007$ to $+0.316$ and $-0.016$ to $+0.250$ at the retrained model. The separation is wide at the official model and narrow at the retrained model.}
\label{fig:appnull}
\end{figure*}

\section{Alternative Accounts and Boundary Tests}
\label{app:alternatives}

The copying interventions establish that a repeated-name-writing ensemble is causally
harmful at the floor. We next test whether several simpler explanations can account for
the observations. The first two concern what produces the harmful repeated-name
preference. The third asks whether the observed attenuation in attention routing requires
an additional learned corrective mechanism.

\subsection{Matching-name amplification}
One possible explanation is that the model detects that the names at $S1$ and $S2$ match
and gives the matching identity an additional boost. Such a mechanism predicts
superadditivity: putting the same name at both subject positions should promote that name
more than the sum of putting it at either position separately.

We test this with a $2\times2$ factorial comparison over whether the identity whose logit is
measured appears at each of the two subject positions. Let $z_{ab}$ denote that identity's
logit when it is present at the first mention if and only if $a=1$, and at the second
mention if and only if $b=1$. The interaction is
\begin{equation}
I = z_{11} - z_{10} - z_{01} + z_{00},
\end{equation}
equivalently $(z_{11}-z_{01})-(z_{10}-z_{00})$, which asks whether placing the identity at
both positions gives a boost beyond the sum of the two single-position effects. A positive
value is superadditive, zero is additive, and a negative value is subadditive. A positive interaction would be the signature predicted
by matching-name amplification. A value near zero would indicate approximately additive
contributions, and a negative value indicates that the two occurrences together contribute
less than the sum of their separate effects.

\begin{figure}[htbp]
\centering
\includegraphics[width=\columnwidth]{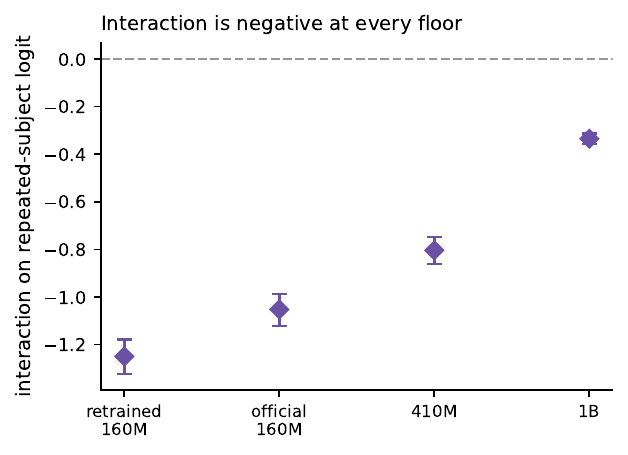}
\caption{Interaction term for the repeated-subject logit at each floor setting. An amplification mechanism predicts a positive value.}
\label{fig:appeq}
\end{figure}

The measured interaction is negative at every floor. It is $-1.250$ with interval
$[-1.324, -1.179]$ for the separately trained 160M model, $-1.052$ with $[-1.122, -0.986]$
for released 160M, $-0.803$ with $[-0.860, -0.750]$ at 410M, and $-0.335$ with
$[-0.356, -0.313]$ at 1B. The tested account therefore predicts the wrong sign, and
repetition is subadditive under this measurement rather than superadditive. Figure~\ref{fig:appeq} plots the interaction at each floor.

This does not mean the model contains no information about whether a name has appeared
before. It rules out the more specific account tested here, in which matching the two
subject identities produces an additional positive boost to the repeated name.

\subsection{A linearly decoded duplication direction}
A second possibility is that the harmful floor behaviour is carried by a single direction
in the residual-stream activation at $S2$ that linearly encodes whether the name has
appeared before.

We fit this direction to repeated and non-repeated second-mention activations from the same 300-prompt 160M floor evaluation used for this control, so it is fitted in distribution rather than held out. The direction is linearly decodable,
so information separating repeated from non-repeated cases is present in the activation.
Decodability alone, however, does not show that this direction causes the harmful
prediction.

We test causality by progressively subtracting the decoded direction from the $S2$
activation at the 160M floor. If this direction were the harmful quantity responsible for
the below-chance preference, removing more of it should improve behaviour.

The opposite occurs. Forced-choice accuracy falls from $0.317$ in the unmodified model to
$0.297$, $0.290$, $0.270$, and $0.230$ as one half, one, two, and four times the decoded
direction are removed. The logit difference likewise becomes more negative at every step,
from $-0.753$ to $-0.807$, $-0.859$, $-0.954$, and $-1.072$.

\begin{figure}[htbp]
\centering
\includegraphics[width=\columnwidth]{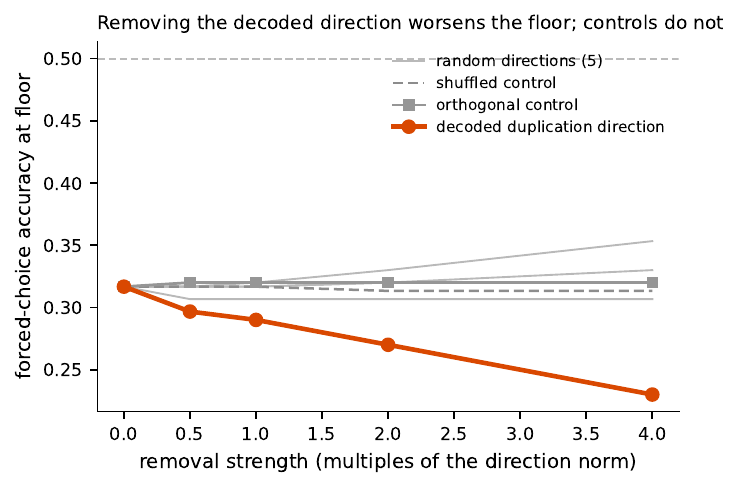}
\caption{Removing the linearly decoded duplication direction at the 160M floor degrades accuracy in proportion to the removal strength.}
\label{fig:appproj}
\end{figure}

Control directions do not produce the same monotonic deterioration. We compare this intervention with three controls evaluated by the same procedure. The
shuffled-label control refits the same linear decoder after randomly permuting the
repeated and non-repeated labels, which removes the relationship between the
representation and the target label. The orthogonal control uses a random direction with
its component along the decoded duplication direction projected out. We also evaluate
five independently sampled random directions. Every direction, including the decoded one,
is scaled to unit norm before removal, so the removal-strength multiplier means the same
thing in every arm. Figure~\ref{fig:appproj} plots the full sweep: the orthogonal,
shuffled and random controls all leave floor behaviour approximately unchanged.

The decoded direction contains information that distinguishes repeated from non-repeated states, but removing it consistently worsens floor behaviour. It is therefore not the harmful causal quantity identified by the copying interventions. This result illustrates why linear decodability and causal importance must be evaluated separately.

\subsection{Testing for an Additional Learned Routing Correction}
The source-identity experiments show that attention heads copy name information from
earlier positions. When the same name occurs at both $S1$ and $S2$, however, some
attenuation of that copying can arise mechanically, because a fixed amount of attention is
distributed across two occurrences of the same identity. We therefore ask whether native
routing does anything measurably more protective than this passive effect.

\begin{figure*}[htbp]
\centering
\includegraphics[width=\textwidth]{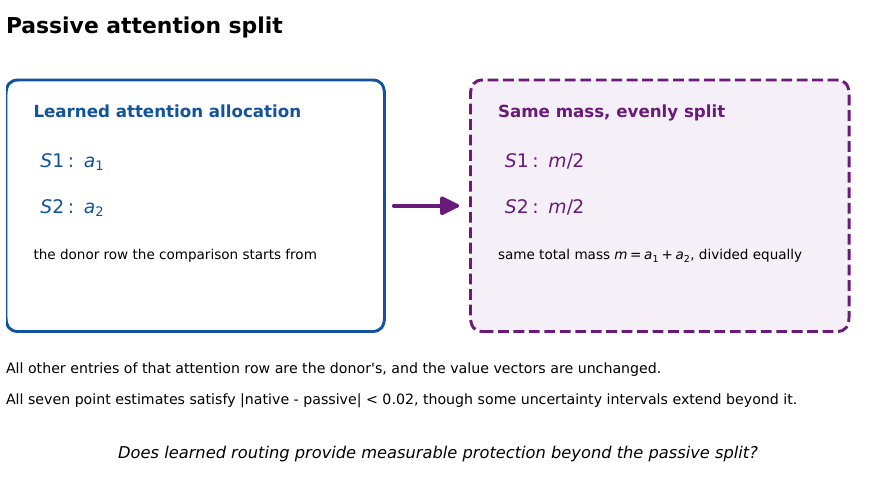}
\caption{Passive attention split. The total attention a copying head places on the two subject mentions is preserved and divided equally between them, leaving the value vectors untouched. This is neither an ablation nor a value replacement, since the same mass is present and only its allocation changes. The donor attention row is taken, the mass it places on the two subject positions is divided equally between them, and the modified row is written back; every other entry of that row is the donor's. All seven point estimates satisfy an absolute difference below $0.02$, although some uncertainty intervals extend beyond it, so this is a small-effect comparison rather than a demonstration of equivalence.}
\label{fig:atlasJ}
\end{figure*}

For each copying head we construct a passive-split comparison. We preserve the total
attention mass the head assigns to $S1$ and $S2$ but divide that mass equally between the
two positions, leaving the value vectors unchanged. This creates a reference in which
attention is redistributed between the two occurrences without introducing an additional
content-dependent routing decision. Figure~\ref{fig:atlasJ} shows the same attention mass before and after the split.

We then compare the native-routing effect with the passive-split effect on the three-name
margin, item by item, and report the native effect minus the passive effect. Starting from the same-second-mention donor's unmodified end-query attention row, we preserve its total mass on the two subject positions but divide that mass equally between them, leaving the rest of the row and the value vectors unchanged. We refer to the unmodified donor allocation as native routing and the equalised version as the passive split. Under this
convention a negative value means native routing is more protective than the passive
split, which is the direction expected if an additional learned routing computation is
actively reducing harmful repeated-name copying.

\begin{figure}[htbp]
\centering
\includegraphics[width=\columnwidth]{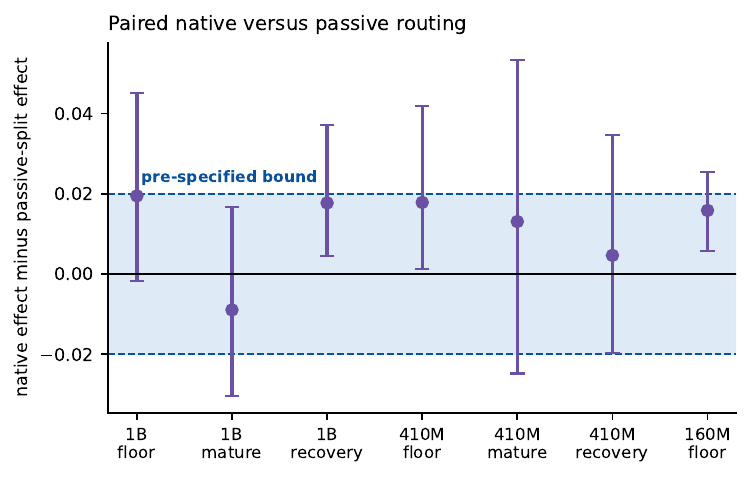}
\caption{Paired differences between the native-routing effect and the passive-split effect, with the analysis bound shaded. Negative values would indicate native routing is more protective than the mechanical split.}
\label{fig:apppassive}
\end{figure}

Across the seven settings for which the comparison is available, every difference has
magnitude below the analysis bound of $0.02$. Six of the seven are positive, the
opposite direction from the prediction above, and several of the template-aware intervals
include zero. Table~\ref{tab:passivefull} gives all seven with both uncertainty procedures.
The separately trained 160M setting has no passive-split arm and is not included. Figure~\ref{fig:apppassive} plots the seven paired differences.

We compare the paired native-minus-passive effect with an analysis bound of 0.02. Every available point estimate lies within this bound, although several template-aware intervals include values outside it. The result therefore constrains the measured difference to be small relative to this comparison value without establishing exact equivalence.

We therefore do not invoke an additional learned routing correction to explain the early
attenuation. The account used in this paper requires repeated-name copying to be harmful
at the floor and mature suppression to become important later. It does not require a
separate early corrective router.

\subsection{Results of the alternative-account tests}
These controls constrain several alternative explanations for the floor behaviour. The repeated-name preference is inconsistent with the tested superadditive amplification account because the measured interaction has the opposite sign. A linearly decoded duplication direction is present, but removing it worsens rather than repairs the prediction. Native attention routing also provides no measurable protection beyond the passive split at the resolution of the comparison.

The positive evidence for copying remains the intervention evidence in Appendix~\ref{app:copying}. Replacing the selected heads improves floor behaviour, and changing the identity supplied at their source changes the identity they support downstream. The resulting account requires harmful repeated-name copying at the floor and later development of mature suppression without requiring a separate early corrective router.

\begin{table*}[t]
\centering\small
\begin{tabular}{@{}lccc@{}}
\toprule
Setting & Mean & Item interval & Nested interval \\
\midrule
160M floor & $+0.016$ & $[0.009,0.023]$ & $[0.006,0.026]$ \\
410M floor & $+0.018$ & $[0.012,0.024]$ & $[0.001,0.042]$ \\
410M recovery & $+0.005$ & $[-0.003,0.012]$ & $[-0.020,0.035]$ \\
410M mature & $+0.013$ & $[0.003,0.023]$ & $[-0.025,0.053]$ \\
1B floor & $+0.019$ & $[0.014,0.026]$ & $[-0.002,0.045]$ \\
1B recovery & $+0.018$ & $[0.012,0.023]$ & $[0.004,0.037]$ \\
1B mature & $-0.009$ & $[-0.015,-0.003]$ & $[-0.031,0.017]$ \\
\bottomrule
\end{tabular}
\caption{Paired difference between the native-routing effect and the passive-split effect on the three-name margin. A negative value means native routing is more protective than the mechanical split, which is the direction an additional learned correction would predict. Six of the seven settings are positive, and all seven are smaller in size than the analysis bound of $0.02$.}
\label{tab:passivefull}
\end{table*}

\section{Identifying and Tracing the Mature Suppressing Head}
\label{app:suppid}

Section~\ref{sec:suppression} follows one attention head backward through training at each
model size. This appendix separates three operations that should not be conflated:
selecting the head, characterising its native computation, and testing it causally.

\subsection{Selection by mature replacement effect}
At the mature checkpoint we evaluate every attention head individually. For a given head we
replace its output vector with that head's mean over the prompts and token positions of
each sentence-template block, leaving the rest of the model unchanged, and measure the change in the $IO$
versus $S$ logit difference. The head whose replacement causes the largest decrease is
selected. This returns L8H9 at 160M, L12H12 at 410M, and L11H0 at 1B.

The selection criterion is therefore behavioural and intervention-based. It asks which
individual head makes the largest measured contribution to the mature model's preference
for the correct name over the repeated name.

The replacement used here differs from the copying intervention in
Appendix~\ref{app:copying}. That experiment constructs one fixed vector from separate
non-repeated control prompts. Here each candidate head is replaced by its own mean over the
prompts and token positions of each sentence-template block. We therefore call this mature-head mean replacement
rather than control-mean replacement.

Because the replacement effect is used to select the head, the selected head's large mature
replacement effect is selection-conditioned. The later causal experiments, including the
parameter transplant and the activation graft, test different quantities and provide
independent evidence.

\subsection{Native attention and direct output projection}
After selecting the head we inspect its behaviour in the unmodified mature model. First we
measure the share of its attention from END that falls on the second occurrence of the
repeated name. Second we compute the direct projection of the head's output at END onto
that name's unembedding direction, where a negative value means the head's own output is
aligned with lowering the repeated name.

These are descriptive attributions. They are not intervention effects and are not used to
select the head.

At maturity the selected head at 160M places $0.934$ of its attention on the second mention
with a direct output projection of $-6.220$. At 410M the values are $0.336$ and $-2.105$,
and at 1B they are $0.527$ and $-7.269$. All three attend to the repeated mention and write
in a direction that lowers it. This characterises what the selected heads do, and should
not be presented as independent confirmation of the selection procedure.

\subsection{Development across training}
We trace the same mature-selected head backward through earlier checkpoints without
reselecting at each step.

At 160M both properties are small around the floor. At step 512 the head places $0.047$ of
its attention on the second mention with projection $-0.005$; at step 1000, $0.021$ and
$-0.016$; and at the floor at step 2000, $0.009$ and $-0.034$. At step 3000, during
behavioural recovery, these change sharply to $0.849$ and $-1.338$. At step 5000 they are
$0.692$ and $-4.590$, and at maturity $0.934$ and $-6.220$.

The trajectory is not globally monotonic. Attention falls slightly across the earliest
checkpoints, rises sharply around recovery, and changes again afterwards. The supported
statement is therefore that the selected head is nearly inactive at the floor and becomes
strongly engaged around recovery, not that either quantity increases monotonically
throughout training.

\begin{figure*}[t]
\centering
\includegraphics[width=0.88\textwidth]{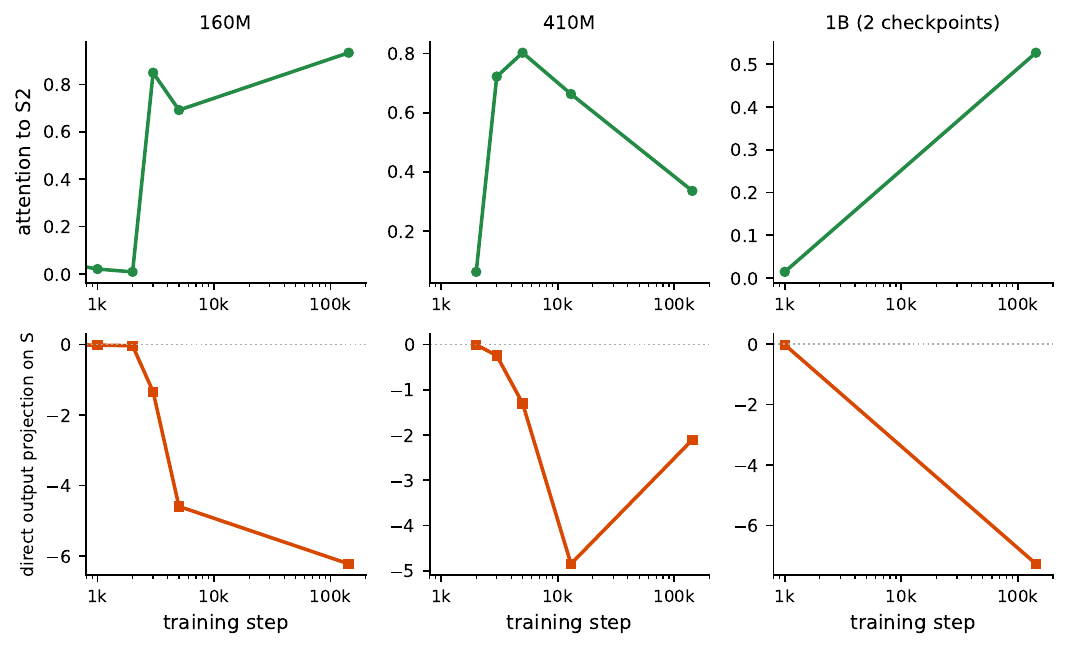}
\caption{The head identified at maturity, traced backward at all three sizes. The upper row shows attention from the final token to the second subject mention and the lower row shows the effect of its output on the repeated-subject logit.}
\label{fig:apptraj}
\end{figure*}

The 410M trajectory shows the same broad separation between a weak floor state and a
strongly engaged later state, and is likewise non-monotonic after recovery. At the floor
attention is $0.062$ and the projection is $-0.005$; at step 3000 they are $0.722$ and
$-0.246$; and at maturity $0.336$ and $-2.105$. For 1B only the floor and mature endpoints
are available, with attention rising from $0.014$ to $0.527$. Figure~\ref{fig:apptraj} and Table~\ref{tab:timing} give the full trajectory. The projection changes
from $-0.025$ to $-7.269$. That supports an endpoint contrast rather than a detailed
trajectory.

\subsection{Suppression develops around behavioural recovery}
The backward trace shows that the computation associated with lowering the repeated name at maturity is weak at the floor and becomes strongly expressed around behavioural recovery. Attention and direct output projection describe the head’s native computation across training. Appendix~\ref{app:suppq} tests its causal importance through output replacement, activation grafting, and parameter transplantation.

\section{Causal Tests of the Mature Suppressing Head}
\label{app:suppq}

This appendix asks whether the mature-selected head contributes to mature performance,
whether its learned computation is sufficient to improve the floor, whether that
improvement is specific, and how much of mature behaviour can be reconstructed when other
helpful heads are added.

\subsection{Contribution to mature behaviour}
We first measure how much mature behaviour depends on the selected head, using the
mature-head mean replacement defined above. At 160M the native mature logit difference is
$3.790$; replacing L8H9 lowers it to $1.131$, a change of $-2.659$, while forced-choice
accuracy remains $0.777$. At 410M it falls from $3.093$ to $1.337$, a change of $-1.755$,
with accuracy $0.900$. At 1B it falls from $3.226$ to $1.634$, a change of $-1.592$, with
accuracy $0.993$.

Replacing the selected head substantially weakens the mature preference while forced-choice accuracy remains above chance at all three sizes. The head is therefore an important part of the recovered computation, while other components also contribute.

\subsection{Backward parameter transplantation}
We next test whether what the head learns by maturity is sufficient to improve the floor.
For each model we copy the mature head's query, key, value, and output matrices into the
same head at the floor checkpoint, together with the head-specific query, key, and value
biases. The attention-layer output bias is shared across heads and is not transplanted.
All other floor parameters remain unchanged.

\begin{figure*}[htbp]
\centering
\includegraphics[width=\textwidth]{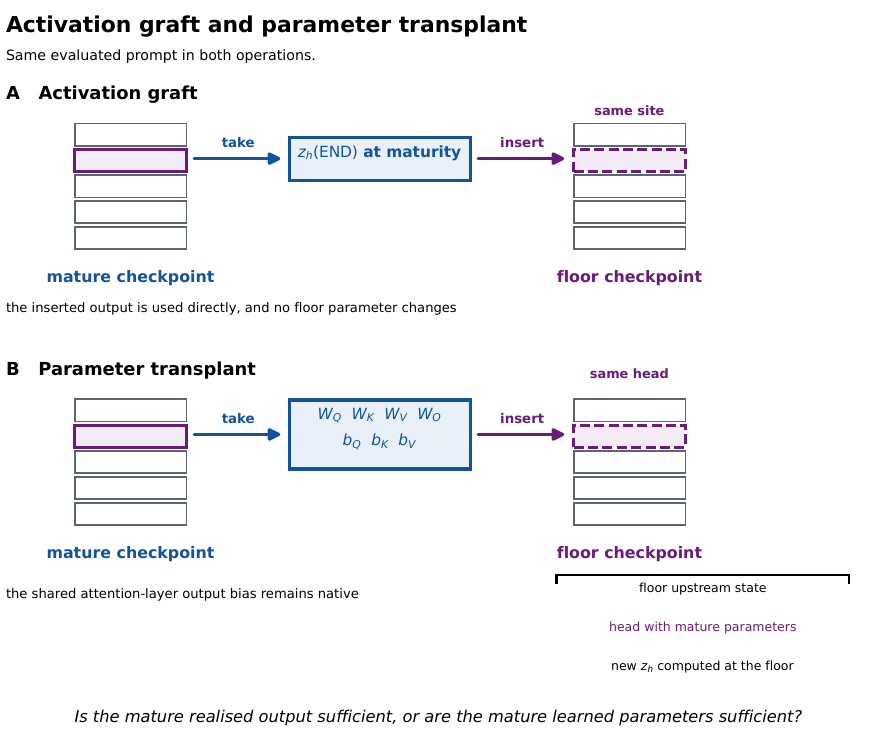}
\caption{Activation graft and parameter transplant. Both move something from the mature checkpoint into the floor model, but they move different objects. The graft copies the head's realised output on the evaluated prompt, so the floor model does not recompute it. The transplant copies the head's query, key, value and output weights together with its head-specific biases, after which the floor model runs normally and the head computes a fresh output from its own upstream representations. The graft asks whether the mature output is sufficient on these prompts; the transplant asks whether what the head has learned is sufficient inside the early network.}
\label{fig:atlasLM}
\end{figure*}

The modified floor model then performs a normal forward pass and recomputes the
transplanted head's activations from its own inputs. This distinguishes parameter
transplantation from activation grafting, in which a precomputed activation is inserted
directly. Figure~\ref{fig:atlasLM} shows what each operation moves and what the floor model then does with it.

The transplant produces a large partial rescue. At 160M the logit difference moves from
$-0.753$ to $+0.849$ with accuracy $0.687$, closing $35.3$ percent of the floor-to-mature
gap. At 410M it moves from $-0.702$ to $+1.876$ with accuracy $0.893$, closing $67.9$
percent. At 1B it moves from $-0.543$ to $+0.984$ with accuracy $0.810$, closing $40.5$
percent.

The mature parameters are therefore sufficient to move every floor checkpoint from a
negative to a positive logit difference. The rescue is substantial, incomplete, and varies
across scale. It supports partial sufficiency, not the claim that one head reconstructs the
entire mature computation. This is stronger than observing that the head's attention or
projection grows during training, because the transplant changes the early model's
behaviour by supplying later-learned parameters.

\subsection{Specificity controls}
\paragraph{Random parameter bundle.} The same parameter-copying operation is applied to a
seeded bundle of eight heads, excluding the selected suppressing head and the twenty heads
with the strongest direct contribution toward the correct name. The bundle closes $-4.7$
percent of the gap at 160M, $10.7$ percent at 410M, and $5.9$ percent at 1B, against
$35.3$, $67.9$, and $40.5$ percent for the single selected head.

\begin{figure}[htbp]
\centering
\includegraphics[width=\columnwidth]{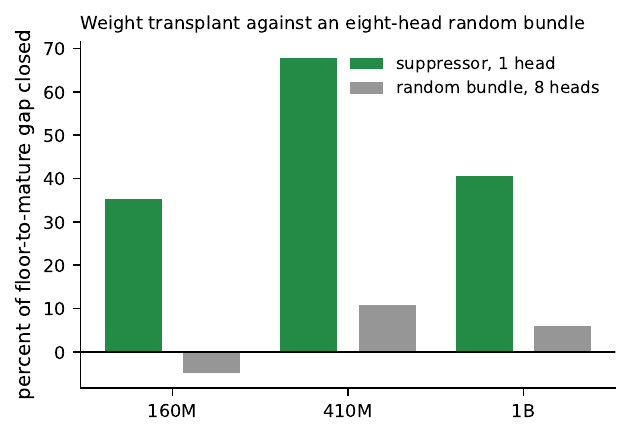}
\caption{Parameter transplant of the single suppressing head against a bundle of eight randomly chosen heads. The comparison is asymmetric in the number of heads transplanted.}
\label{fig:appbundle}
\end{figure}

This comparison favours the control in parameter count because the control receives eight heads while the treatment receives one. It shows that copying mature
parameters into the floor is not generically sufficient to produce the observed rescue. It
is not a head-for-head matched comparison and should not be described as one. Figure~\ref{fig:appbundle} compares the two transplants.

\paragraph{Output-norm-matched activation graft.} A separate experiment supplies the mature
head's activation rather than its parameters. For each prompt the mature activation of the
selected head is inserted at the corresponding site in the floor model, so the floor model
does not recompute the head from its own query, key, and value inputs. A control head whose output norm is closest to the
suppressor's is chosen, and the same operation is performed with it.

At 160M, grafting the suppressing head changes the logit difference by $+0.583$ against
$+0.013$ for the output-norm-matched control head. At 410M the values are $+0.752$ and $-0.351$.

\begin{figure}[htbp]
\centering
\includegraphics[width=\columnwidth]{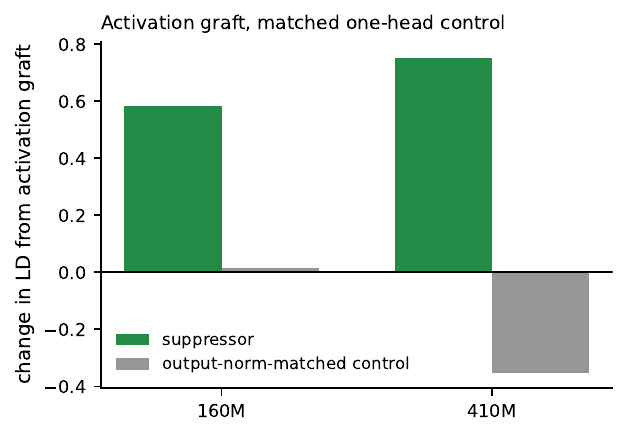}
\caption{Injecting the mature head's activations into the floor checkpoint, with an output-norm-matched control head as control.}
\label{fig:appgraft}
\end{figure}

This is a one-for-one specificity control, but it answers a different causal question from parameter transplantation. The graft asks whether the mature head's realised output on these prompts
can improve the floor. The transplant asks whether the mature parameters can recompute a
helpful output inside the early network. The two are convergent evidence rather than
results to be pooled. Figure~\ref{fig:appgraft} plots both grafts.

\subsection{Reconstruction with correct-name writers}
The one-head transplant leaves part of the mature gap unclosed, which is expected because
mature behaviour also depends on components that promote the correct name.

At 410M the selected head alone closes $67.9$ percent of the gap. Adding the mature
parameters of the strongest correct-name-writing heads cumulatively raises this to $88.1$
percent with all eight writers added, at forced-choice accuracy $0.933$.

\begin{figure*}[htbp]
\centering
\includegraphics[width=\textwidth]{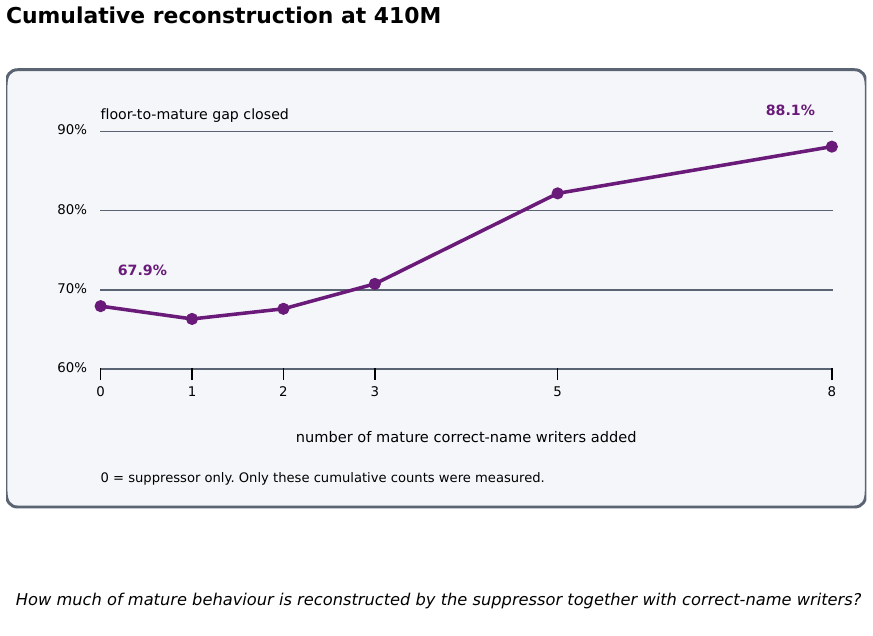}
\caption{Cumulative reconstruction at 410M. Mature parameters are transplanted into the floor model cumulatively, beginning with the suppressing head alone and then adding the strongest correct-name writers in order. The result covers selected cumulative prefixes rather than every intermediate count: measurements exist at zero, one, two, three, five and eight writers. The suppressing head alone closes $67.9$ percent of the floor-to-mature gap and all eight writers together reach $88.1$ percent, at forced-choice accuracy $0.933$. The experiment is available at this size only.}
\label{fig:atlasO}
\end{figure*}

This supports a division of labour. Figure~\ref{fig:atlasO} plots the gap closed as writers are added.

The suppressing head removes support from the repeated
competitor while correct-name writers promote the correct one, and the mature behaviour is
therefore not reconstructed by the suppressor alone. This experiment is available at 410M and is not a three-scale result.

\subsection{Output scaling leaves the floor below chance}

\begin{figure*}[htbp]
\centering
\includegraphics[width=\textwidth]{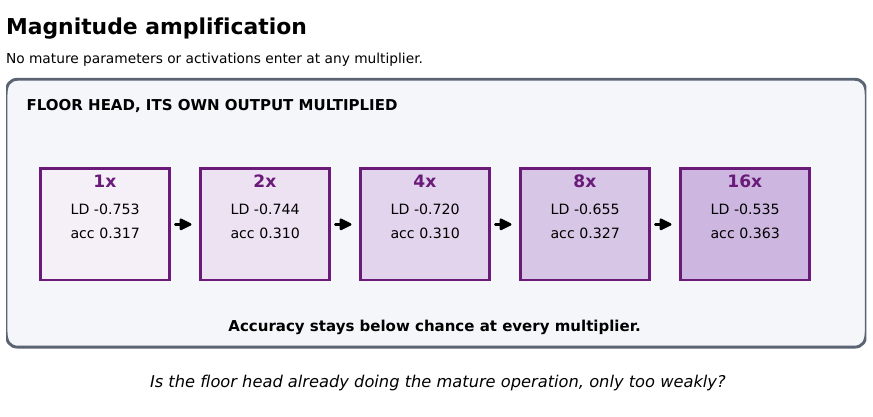}
\caption{Magnitude amplification. The floor head's own output is multiplied by two, four, eight and sixteen, with no mature parameters introduced. If the floor head were already performing the mature operation only weakly, scaling it should move behaviour toward maturity. Accuracy remains below chance at every multiplier, which is why the transplant result is read as supplying a changed computation rather than additional strength.}
\label{fig:atlasN}
\end{figure*}

One possible explanation of the transplant result is that the floor head already performs
the mature operation but too weakly. If so, multiplying the floor head's own output should
improve behaviour. It does not.

At 160M, scaling the immature head's output by two, four, eight, and sixteen moves the
logit difference from $-0.753$ to $-0.744$, $-0.720$, $-0.655$, and $-0.535$, with accuracy
moving from $0.317$ to $0.363$ at the largest factor. The floor remains below chance
throughout. At 410M the corresponding sixteen-fold values are $-0.590$ and $0.327$, again
below chance.

The floor head is therefore not a smaller version of the mature suppressing computation. Figure~\ref{fig:atlasN} gives the logit difference and accuracy at each multiplier.

The transplant supplies a change in what the head computes, including its routing and
output transformation, rather than only an increase in magnitude.

\subsection{The mature suppressing head contributes and partially rescues the floor}
Four conclusions follow. The selected head makes a substantial contribution to the
magnitude of mature correctness, but replacing it does not reproduce the floor. Its mature
parameters are partially sufficient to rescue the floor at all three primary sizes. Both
the eight-head parameter control and the matched one-head activation control show the
rescue is not a generic consequence of importing mature components. And amplifying the
immature version fails, so the critical developmental change concerns the head's learned
computation rather than its raw magnitude.

Together with the trajectory above, these results support a mature suppressing computation
that is weak during the below-chance window, develops around recovery, and then contributes
causally to the recovered behaviour.

\begin{figure}[htbp]
\centering
\includegraphics[width=\columnwidth]{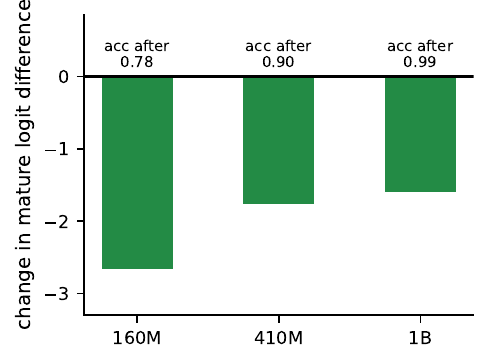}
\caption{Replacing the suppressing head's prompt-specific output with its mature-head mean, taken over the prompts and token positions of each sentence-template block. This is not the control mean of Appendix~\ref{app:copying}. The logit difference falls substantially at all three sizes while forced-choice accuracy remains well above chance, so the head is a substantial contributor to the mature preference rather than solely responsible for it.}
\label{fig:appablation}
\end{figure}

\section{Distributed Upstream Dependence of the Suppressing Head}
\label{app:upstream}

The mature suppressing head depends on an early feed-forward-block state, while the direct
connection from that block to the head carries almost none of the measured dependence.

\subsection{Early feed-forward-block replacement}
For each prompt we construct a matched non-repeated control prompt on the same sentence
template, run it, and cache the output of its first feed-forward block. We then run the
original repeated-name prompt while replacing only that block's output with the cached
control activation. The original tokens are unchanged, so the repeated name remains present
at the second mention and any downstream change comes from replacing the residual-stream activation
rather than editing the input.

Across the tested early blocks, the first produces the largest behavioural change.
Replacing its output with the matched non-repeated value substantially reduces the harmful
floor preference and can bring forced-choice accuracy close to chance.

This identifies the first feed-forward block as an important early source of repetition-sensitive state. Because the intervention replaces the block’s complete output, it identifies the source of the causal effect without isolating a single feature within that state.

\subsection{Dependence of the mature suppressing head}
At maturity we apply the first-block replacement and then measure the selected head's
attention to the second mention. It decreases from $0.93$ to $0.68$ at 160M, from $0.34$ to
$0.16$ at 410M, and from $0.53$ to $0.28$ at 1B.

The decrease occurs at all three sizes, showing that the mature suppressing head's routing
depends partly on information computed early in the network. The dependency is incomplete,
since substantial attention remains after replacement. This supports a causal dependency
from the early state into the mature head's behaviour, but does not identify the route by
which it is transmitted.

\subsection{Isolating the direct route}
A head's query input contains the sum of many earlier residual contributions, so replacing
the entire query input would conflate the first block's contribution with every other
upstream one.

\begin{figure*}[htbp]
\centering
\includegraphics[width=\textwidth]{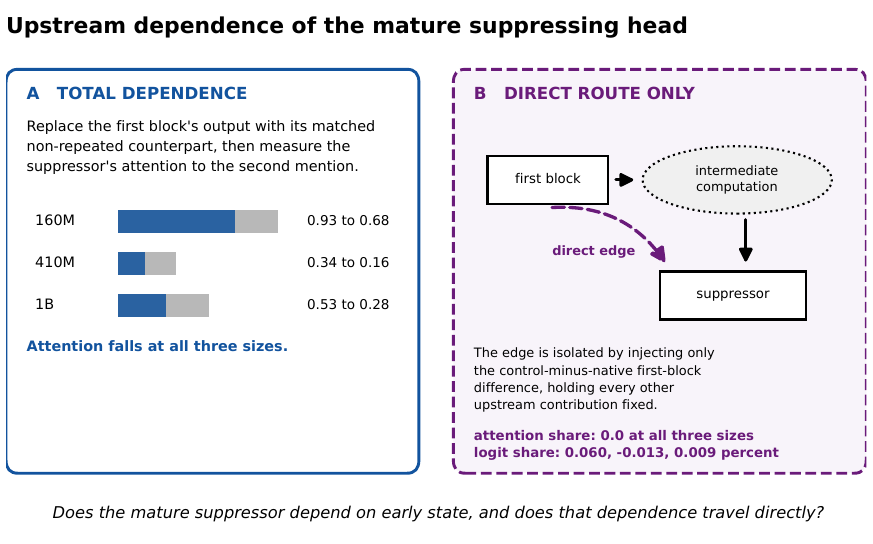}
\caption{Upstream dependence of the mature suppressing head. Panel A replaces the first feed-forward block's output with its matched non-repeated counterpart and measures the suppressor's attention to the second mention, which falls at all three sizes. Panel B isolates only the direct connection between the two, adding the first-block difference to the suppressor's downstream input while holding every other upstream contribution fixed; that route carries none of the attention dependence and well under a tenth of a percent of the logit dependence. The influence therefore reaches the head through intermediate computation, drawn as an unresolved region because the experiments do not identify which components form the path.}
\label{fig:atlasPQ}
\end{figure*}

To isolate only the direct route, we compute the difference between the first block's
output on the matched control and on the original prompt, and add only that difference to
the suppressing head's downstream input, leaving every other residual contribution
unchanged. This edge-level intervention asks how much of the total first-block dependence
reaches the head directly, without passing through intermediate components. Figure~\ref{fig:atlasPQ} separates the total first-block dependence from the isolated direct route.

Two shares are measured. On the head's attention to the second mention, the direct route
carries exactly $0.0$ at all three sizes. On the final logit difference it carries $0.060$
percent at 160M, $-0.013$ percent at 410M, and $0.009$ percent at 1B. Both are effectively
zero, and the small negative logit estimate at 410M should be read as estimation variation
around zero rather than a meaningful negative route. We report both because the dependence
described above is measured on attention while the routing literature more often reports a
logit share, and the two agree here.

Because the head measurably depends on the first-block state while the isolated direct
route carries approximately none of that dependence, the information must reach the head
through intermediate computation.

\subsection{Upstream dependence passes through intermediate computation}
The experiments establish that replacing the first feed-forward block's state changes floor
behaviour, that the mature suppressing head partly depends on that state, and that the
direct route from the block to the head carries approximately none of the measured
dependence.

The appropriate conclusion is distributed upstream dependence rather than a single direct
connection from the first feed-forward block to the suppressing head. Intermediate
attention heads, feed-forward blocks, and residual-stream paths may all contribute, but
this appendix does not assign the dependence to a particular sequence.

\begin{table}[htbp]
\centering\footnotesize
\setlength{\tabcolsep}{3pt}
\begin{tabular}{@{}lcc@{}}
\toprule
Checkpoint & \makecell{Attention\\to $S2$} & \makecell{Direct output\\projection on $S$} \\
\midrule
step 512 & $0.047$ & $-0.005$ \\
step 1000 & $0.021$ & $-0.016$ \\
step 2000 (floor) & $0.009$ & $-0.034$ \\
step 3000 & $0.849$ & $-1.338$ \\
step 5000 & $0.692$ & $-4.590$ \\
step 143000 (mature) & $0.934$ & $-6.220$ \\
\bottomrule
\end{tabular}
\caption{The mature-defined suppressing head traced backward at 160M. It is barely engaged by these diagnostics at the floor; they are descriptive, not causal.}
\label{tab:timing}
\end{table}

\section{Extensions Across Scale and Model Family}
\label{app:scale}

The detailed copying-to-suppression account is established on the released Pythia 160M,
410M, and 1B models. This appendix examines which parts of that account extend to
additional Pythia sizes and to independently trained GPT-2 and OLMo models.

The interventions differ across the settings below. We therefore interpret the sign and
timing within each experiment rather than comparing effect magnitudes across settings.

\subsection{Additional Pythia sizes}
\subsubsection{First feed-forward block replacement across five sizes}
\begin{figure*}[htbp]
\centering
\includegraphics[width=\textwidth]{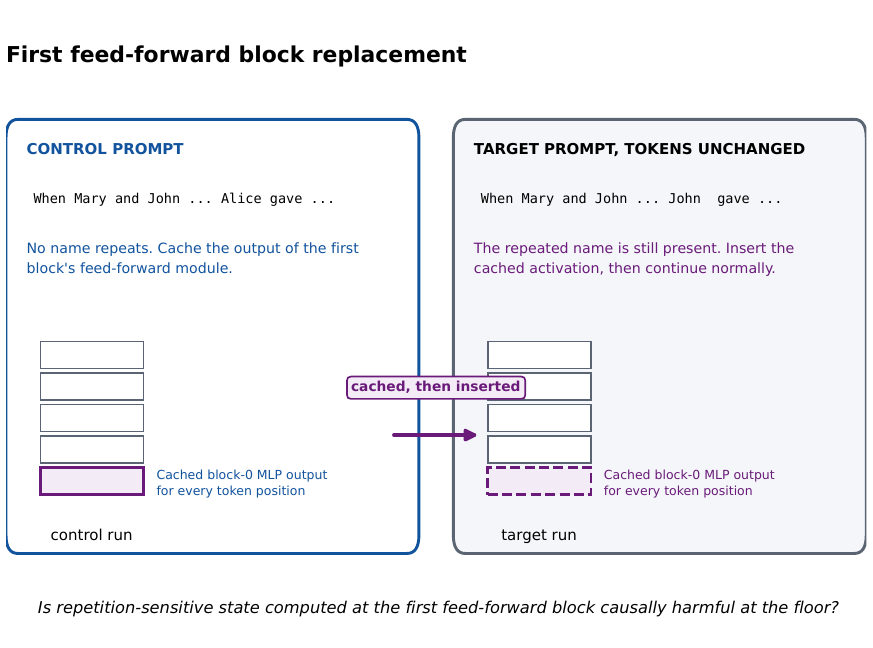}
\caption{First feed-forward block replacement, also called the layer-0 MLP. The evaluated prompt is not edited. A matched control prompt in which no name repeats is run first, the output of that one module is cached, and the cached activation is inserted at the same site in the target run before the forward pass continues. Because the whole module output is replaced, the experiment does not isolate a duplication-specific feature: it establishes the first feed-forward block as an important early source of repetition-sensitive state rather than as a duplication detector.}
\label{fig:atlasF}
\end{figure*}

For each prompt we construct a matched control on the same sentence template with the same
correct name, in which the second occurrence of the repeated name is replaced by a
different name so that no name appears twice. We run the model on the control prompt and
cache the output of the first transformer block's feed-forward module, then run the
original repeated-name prompt while replacing that module output with the cached control
activation. The evaluated prompt's tokens are unchanged, so this is an internal activation
replacement rather than an input edit. Figure~\ref{fig:atlasF} follows the cached activation from the control run into the target run.

\begin{table}[htbp]
\centering\small
\setlength{\tabcolsep}{4pt}
\begin{tabular}{@{}lccc@{}}
\toprule
Model & Native floor $\ld$ & Change & 95\% interval \\
\midrule
Pythia 70M & $-2.327$ & $+1.808$ & \footnotesize$[1.694, 1.926]$ \\
Pythia 160M & $-0.757$ & $+0.769$ & \footnotesize$[0.687, 0.854]$ \\
Pythia 410M & $-0.616$ & $+0.149$ & \footnotesize$[0.101, 0.198]$ \\
Pythia 1B & $-0.651$ & $+0.493$ & \footnotesize$[0.449, 0.539]$ \\
Pythia 1.4B & $-0.517$ & $+0.371$ & \footnotesize$[0.338, 0.406]$ \\
\bottomrule
\end{tabular}
\caption{Change in the floor logit difference when the first feed-forward block's output is replaced by its matched non-repeated counterpart. The effect is positive at all five sizes and every interval lies above zero.}
\label{tab:fivescale}
\end{table}

Table~\ref{tab:fivescale} gives the change at each size, and replacing an early activation
with its non-repeated counterpart improves floor behaviour across the whole range. The effect is much smaller at 410M than elsewhere, and we retain that heterogeneity
rather than presenting the result as a uniform scaling law.

Across these five sizes, the first feed-forward block carries repetition-sensitive information that harms the floor prediction. The detailed head-level copying analysis is evaluated separately in the three primary models.

\subsubsection{Pythia 2.8B}
We report two distinct interventions at this size and keep their interpretations separate.

\paragraph{Input edit.} The second occurrence of the repeated name is replaced directly in
the token sequence by a third name. At the floor the unmodified model has forced-choice
accuracy $0.297$ and logit difference $-0.504$, and this edit improves the logit difference
by $+0.456$. At maturity the same edit changes it by $-4.076$, so input-level sensitivity to
repetition reverses between the floor and maturity.

\paragraph{Early feed-forward activation replacement.} A separate experiment replaces the
outputs of the early feed-forward blocks while leaving the prompt tokens unchanged.
Resampling those activations from matched non-repeated prompts raises floor accuracy from
$0.297$ to $0.603$. Replacing the same activations with their mean instead gives $0.467$.
The replacement distribution therefore matters, and we report the two separately rather
than treating feed-forward replacement as a single intervention.

At 2.8B, both the token edit and the internal activation replacements show harmful early sensitivity to the repeated relation. The input-edit effect additionally reverses at maturity.

\subsection{Independently trained GPT-2 models}
We evaluate two independently trained GPT-2 Small models released by the Stanford CRFM
Mistral project \citep{karamcheti2021mistral}, \texttt{alias-gpt2-small-x21} and
\texttt{battlestar-gpt2-small-x49}. For consistency with the paper-wide convention, all intervention effects in this subsection are reported as replaced minus native.

The mature-defined suppressing head is L10H10 in the first run and L10H7 in the second.
Each is fixed at the mature checkpoint by its replacement effect and then traced backward
through training. Their differing identities already indicate that the developmental role
is more stable than the component implementing it.

We apply two interventions. The first replaces the output of the first feed-forward block
with the corresponding output from a matched non-repeated prompt, following the procedure
above. The second applies mature-head mean replacement to the head that later becomes the
mature suppressor, measuring how much the model at each checkpoint depends on that future
component.

At step 1500 the first run has native logit difference $-1.475$, and the early activation
replacement changes it by $+1.123$ while replacing the future suppressor changes it by
about $-0.002$. The second run has native logit difference $-1.741$, with corresponding
values of $+1.472$ and about $-0.002$.

At maturity both interventions reverse. The early activation replacement changes the logit
difference by $-4.081$ and $-4.607$ in the two runs. Replacing the mature suppressing head
changes it by $-1.928$ and $-1.506$, lowering the mature logit difference from $3.408$ to
$1.480$ and from $3.647$ to $2.142$. Every value quoted here is the effect of the
intervention, so a negative number means the replacement moved the model away from the
correct name and the native component was therefore helping.

The common result is an ordering. Repetition-sensitive early computation is already
causally important at the poor checkpoint, the head that later performs substantial
suppression is nearly inert there, and that component becomes important later. The
suppressor identity and the detailed timing differ between the two runs, so the
developmental ordering recurs more robustly than the identity of the implementing head.

\subsection{OLMo-1B}
We evaluate OLMo-1B \citep{groeneveld2024olmo} at the early revision
\texttt{step1000-tokens4B} and at the released final revision. At the early checkpoint
forced-choice accuracy is $0.437$ and the logit difference is $-0.220$. At maturity
accuracy is $1.000$ and the logit difference is $4.425$.

For each prompt we construct a matched non-repeated control on the same sentence template
and replace the residual-stream activation at the second occurrence of the repeated name
with the corresponding activation from that control, across layers 4 to 7. The prompt
tokens and all activations outside the selected position and layer band remain native.

At the early checkpoint the replacement improves the logit difference by $+0.0448$ with
interval $[+0.0349, +0.0549]$. At maturity the same intervention changes it by $-5.137$
with interval $[-5.257, -5.018]$. The causal role of the second-mention residual state
therefore reverses between the two evaluated checkpoints.

The state-level developmental reversal therefore extends beyond Pythia and GPT-2. This experiment evaluates the changing causal role of the second-mention state in OLMo, while the full head-level copying account is evaluated in the primary Pythia models.

\subsection{Cross-scale and cross-family summary}
The extensions above support a recurring developmental motif without implying one universal
circuit. Across additional Pythia sizes, disrupting repetition-sensitive early computation
improves behaviour at the floor. In two independently trained GPT-2 models that early
computation is already causally important while the future mature suppressor remains nearly
inert. In OLMo the causal role of the second-mention residual state reverses between an
early and a mature checkpoint.

The strongest copying-specific evidence remains the donor-free ensemble replacement and the
source-identity intervention on the primary models. The additional settings each establish a
subset of the account without showing that the same heads or the same
intervention-defined mechanism recur everywhere. Table~\ref{tab:crossfamily} summarises every setting.

\begin{table*}[t]
\centering\small
\setlength{\tabcolsep}{4pt}\renewcommand{\arraystretch}{1.2}\footnotesize
\begin{tabular}{@{}>{\raggedright\arraybackslash}p{2.0cm}>{\raggedright\arraybackslash}p{2.3cm}>{\raggedright\arraybackslash}p{3.3cm}>{\raggedright\arraybackslash}p{1.9cm}>{\raggedright\arraybackslash}p{1.9cm}>{\raggedright\arraybackslash}p{3.1cm}@{}}
\toprule
Setting & Early state & Intervention & Early result & Mature result & Strongest supported conclusion \\
\midrule
Pythia 70M to 1.4B & $\ld$ negative at all five floors & first feed-forward block replacement & positive at all five & not evaluated here & early repetition-sensitive computation is harmful across scale \\[2pt]
Pythia 2.8B & acc. $0.297$, $\ld$ $-0.504$ & input edit & $+0.456$ & $-4.076$ & input-level sensitivity to repetition reverses \\[2pt]
GPT-2 alias & $\ld$ $-1.475$ at step 1500 & early-block replacement; future-suppressor replacement & $+1.123$; $-0.002$ & $-4.081$; $-1.928$ & early harmful computation precedes mature suppression \\[2pt]
GPT-2 battlestar & $\ld$ $-1.741$ at step 1500 & as above & $+1.472$; $-0.002$ & $-4.607$; $-1.506$ & ordering recurs with a different suppressor head \\[2pt]
OLMo-1B & acc. $0.437$, $\ld$ $-0.220$ & second-mention residual-state replacement & $+0.0448$ & $-5.137$ & state-level causal role reverses outside Pythia and GPT-2 \\
\bottomrule
\end{tabular}
\caption{Extensions across scale and model family. Every entry is the effect of the intervention on the logit difference, so a positive value means the replacement moved the model toward the correct name and a negative value means the native component was helping. Interventions differ across rows, so their magnitudes are not directly comparable. The final column states only the conclusion supported by the intervention in that row.}
\label{tab:crossfamily}
\end{table*}

\FloatBarrier
\section{Extensions Across Tasks}
\label{app:tasks}

\begin{figure*}[htbp]
\centering
\includegraphics[width=\textwidth]{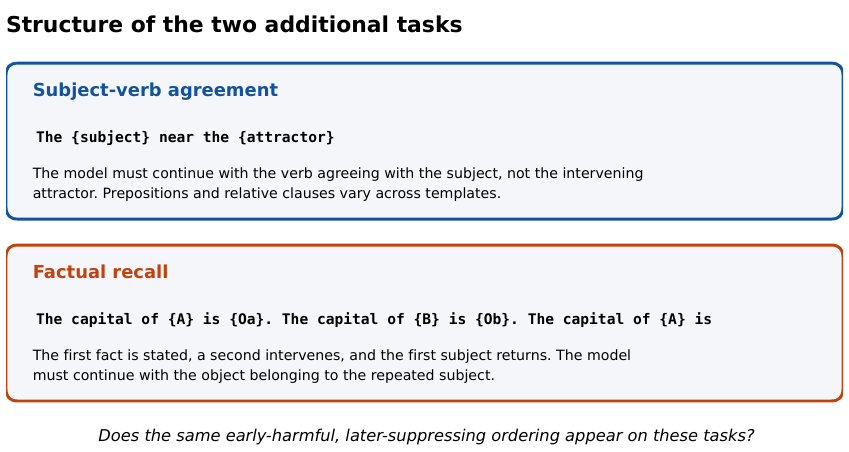}
\caption{Prompt structures for the subject-verb agreement and factual-recall tasks. The subject-verb agreement prompts place an attractor of the opposite number between the subject and the verb position, with the preposition or relative clause varying across templates. The factual-recall prompts state one fact, interpose a second, and then return to the first subject, so the correct continuation is the object belonging to the repeated subject.}
\label{fig:appstructures}
\end{figure*}

The main analysis concerns the naming task. We next examine whether related developmental patterns appear on other structured prediction tasks. Figure~\ref{fig:appstructures} illustrates the subject-verb agreement and factual-recall prompts. Each task uses its own score and intervention, allowing us to compare developmental timing without assuming that the same circuit implements every result.

\subsection{Greater-Than}
\subsubsection{Task and score}
Each prompt describes an event spanning two years and ends partway through the closing
year, for example ``The war lasted from the year 1723 to the year 17''. A plausible
continuation completes the century with a two-digit value greater than 23.

Writing $YY$ for the final two digits of the starting year, we define
\begin{equation}
\mathrm{GT} = \sum_{v > YY} p(v) - \sum_{v \le YY} p(v),
\end{equation}
where $v$ ranges over the valid two-digit continuation tokens. A positive score means the
model assigns more probability to continuations representing a later year. Forced-choice
accuracy is the fraction of prompts with a positive score, so chance is one half.

We use only starting years for which the year tokenises into an unambiguous century token
followed by a two-digit token. The $YY$ position is located directly in each tokenised
prompt rather than inferred from a fixed offset.

\subsubsection{Matched-state intervention}
For every target prompt we construct a matched prompt with the same event description and a
different starting year, so the two differ only in the year. We replace the residual-stream
activation at the target's $YY$ token with the corresponding activation from the matched
prompt across an early layer band, and the rest of the forward pass proceeds normally.

At step 1000 the change in the score is $+0.082$ at 160M with interval
$[0.017, 0.148]$, $+0.097$ at 410M with $[0.036, 0.158]$, and $+0.319$ at 1B with
$[0.285, 0.356]$. At maturity the same intervention changes the score by $-0.737$ with
$[-0.824, -0.651]$, $-0.667$ with $[-0.747, -0.588]$, and $-0.733$ with
$[-0.819, -0.648]$. The intervention effect therefore reverses sign across training at all
three sizes.

At 160M and 1B the early behavioural score is also below chance, at $0.320$ and $0.267$. At
410M early accuracy is already $0.553$. The 410M result is therefore a causal reversal in
the role of the year state without a corresponding below-chance behavioural window. The Greater-Than experiment therefore shows that the causal contribution of a task-relevant state can reverse during training. Such a reversal can occur whether or not the early behavioural score falls below chance.

\subsection{Subject-verb agreement}
This task asks the model to prefer the verb form agreeing with the grammatical subject even when an
opposite-number attractor noun appears between the subject and the verb. The intervention
replaces the residual-stream activation at that attractor noun with the corresponding activation from a matched control prompt.

The result does not follow the developmental pattern studied in this paper. The effect is
positive at every evaluated checkpoint, moving from $+0.020$ with interval
$[0.018, 0.023]$ at step 256 to $+0.022$ with $[0.017, 0.028]$ at maturity. There is no
sign reversal and the effect remains small throughout training.

Subject-verb agreement therefore provides a contrasting case in which the tested intervention remains small and positive throughout training. Changing causal roles are not automatic across structured linguistic tasks.

\subsection{Factual recall with a competing entity}
Each prompt introduces two entities and asks for a property of one of them, so the more
recently mentioned but incorrect entity competes with the correct answer. The behavioural
score is the correct entity's logit minus the competing entity's logit, and forced-choice
accuracy is the fraction of prompts for which this is positive.

Across training, accuracy changes from $0.567$ at step 512 to $0.083$ at step 3000 and then
recovers to $0.917$ at maturity, so this task exhibits a pronounced behavioural dip of its
own.

We identify a task-specific head from the mature checkpoint that contributes to lowering
the competing entity, and trace that same head backward through training. Mature-head mean replacement
changes the score by $-0.029$ at step 2000 and $+0.044$ at step 3000, both close to zero.
At maturity, replacement changes the score by $-1.330$ from a native score of $1.970$,
meaning the native head contributes substantially to the mature preference.

This provides evidence for a developmental pattern in which a mature suppressing component remains weak during poor intermediate behaviour and becomes substantial during recovery. The copying-specific interventions used for the naming task were not run on factual recall.

\subsection{Cross-task summary}
\begin{table*}[t]
\centering\small
\setlength{\tabcolsep}{4pt}\renewcommand{\arraystretch}{1.2}\footnotesize
\begin{tabular}{@{}>{\raggedright\arraybackslash}p{2.1cm}>{\raggedright\arraybackslash}p{2.7cm}>{\raggedright\arraybackslash}p{3.2cm}>{\raggedright\arraybackslash}p{2.0cm}>{\raggedright\arraybackslash}p{1.9cm}>{\raggedright\arraybackslash}p{2.6cm}@{}}
\toprule
Task & Early behaviour & Intervention & Early result & Mature result & Strongest supported conclusion \\
\midrule
Naming & below chance at 160M, 410M and 1B & control-mean copying-head output replacement & $+0.180$ to $+0.804$ & separate suppression experiments & full copying-to-suppression account in the primary models \\[2pt]
Greater-Than & below chance at 160M and 1B; $0.553$ at 410M & residual-state replacement at $YY$ & $+0.082$ to $+0.319$ & $-0.737$ to $-0.667$ & causal role of the year state reverses \\[2pt]
Factual recall & dips to $0.083$ & mature-head mean replacement & near zero through the dip & replacement effect $-1.330$ & behavioural dip with a later substantial suppressing component \\[2pt]
Subject-verb agreement & no corresponding dip established & matched residual-state replacement & $+0.020$ & $+0.022$ & no reversal; negative case \\
\bottomrule
\end{tabular}
\caption{Task extensions. Each row uses a task-specific intervention and outcome, so effect magnitudes should not be compared across tasks. The final column gives the strongest conclusion supported within that task.}
\label{tab:crosstask}
\end{table*}

Across the additional tasks the evidence is deliberately uneven. Greater-Than shows a clear
causal sign reversal in the role of an early token state at all three model sizes, although
only two of them also fall below chance. Factual recall shows a deep behavioural dip
accompanied by a mature-defined suppressing component that remains weak until later in
training. Subject-verb agreement provides a negative case in which the tested intervention
remains small and positive throughout.

These results show that temporary developmental failures and early-to-late changes in
causal role are not unique to the naming task. Table~\ref{tab:crosstask} summarises the four tasks, and the full copying-specific account remains restricted to
the three primary Pythia models.

\FloatBarrier
\section{Training From Scratch and Ordinary-Text Controls}
\label{app:training}

This appendix uses three complementary training experiments to examine how the developmental phenomenon can arise.

Controlled task injection tests how the frequency of structured task examples affects learning and generalisation. A minimal symbolic task tests whether a copy-heavy distribution with a rarer override rule is sufficient to produce a dip and intervention reversal. Models trained only on ordinary text test whether harmful repetition sensitivity can appear without explicit task injection. Each experiment addresses a different part of the training account.

\subsection{Controlled task injection}
\subsubsection{Model and training setup}
We train an eight-layer GPT-2-style transformer with eight attention heads, model dimension
512, and a context length of 256 tokens. The feed-forward width is left at the library
default of four times the model dimension, giving 2048.

Training uses AdamW \citep{loshchilov2019adamw} with learning rate $3\times10^{-4}$, weight
decay $0.01$, betas $0.9$ and $0.95$, batch size 32, a linear warmup over the first 1000
steps, and a cosine decay to one tenth of the peak learning rate, for 40000 steps in total.

The ordinary-text component is drawn from WikiText-103 \citep{merity2017pointer}. At each
batch position an ordinary-text sequence is replaced by a task-structured sequence with
probability equal to the injection rate, and the four rates are $0$, $0.05$, $0.15$, and
$0.30$. The task pool contains 4000 preconstructed sequences assembled from eight sentence
frames, twelve place words, and twelve object words. In the principal comparison the four
conditions share the same parameter initialisation and the same ordinary-text stream, so
the injection rate is the intended difference between them. Additional independently seeded runs test whether the qualitative trajectories persist
under broader training randomness, including model initialisation and the sampling of both
the ordinary text and the injected task rows.

\subsubsection{Held-out and in-distribution evaluations}
The sentence frames are shared between training and evaluation because they define the task
structure. The names are not. Injected sequences draw from one name pool while the held-out
evaluation uses a disjoint pool.

\begin{figure*}[htbp]
\centering
\includegraphics[width=\textwidth]{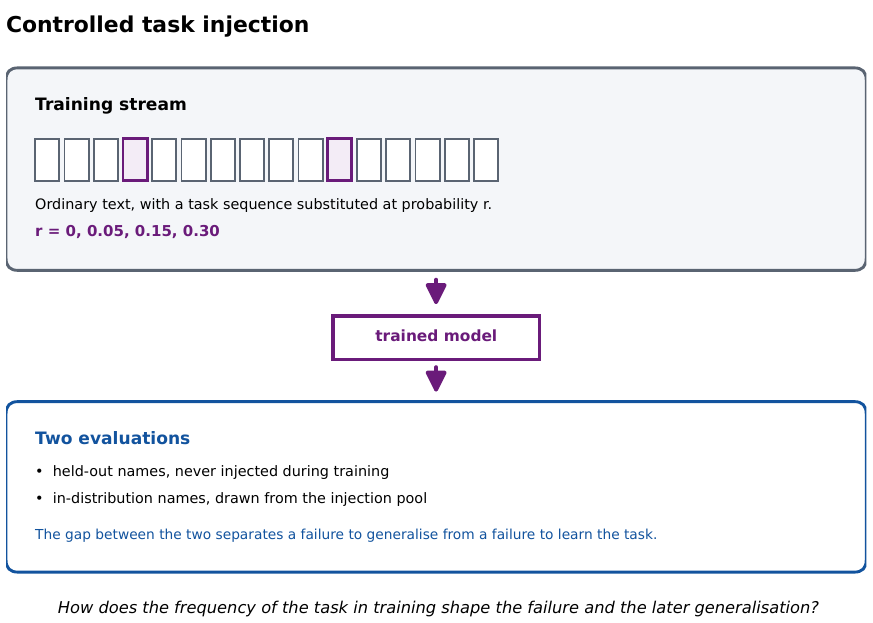}
\caption{Controlled task injection. Models are trained from scratch on ordinary text into which task-structured sequences are substituted at a controlled rate, and each checkpoint is then evaluated twice: on names never injected during training, and on names drawn from the injection pool. The gap between the two separates a failure to generalise to unseen names from a failure to learn the task at all.}
\label{fig:atlasR}
\end{figure*}

Each checkpoint is evaluated on a held-out set of 300 prompts using names absent from the
injected sequences, and on an in-distribution set of 300 prompts using names from the
injection pool. The held-out set measures whether the learned rule generalises to unseen
names. The in-distribution set distinguishes a failure to generalise from a failure to
learn the injected task at all. A model may therefore reach ceiling in distribution while
remaining substantially below ceiling on held-out names. Figure~\ref{fig:atlasR} shows the training mixture and the two evaluations it feeds.

\subsubsection{Results}
With no task injection the model does not acquire the override rule, ending at held-out
accuracy $0.283$ and in-distribution accuracy $0.217$.

\begin{figure*}[t]
\centering
\includegraphics[width=0.88\textwidth]{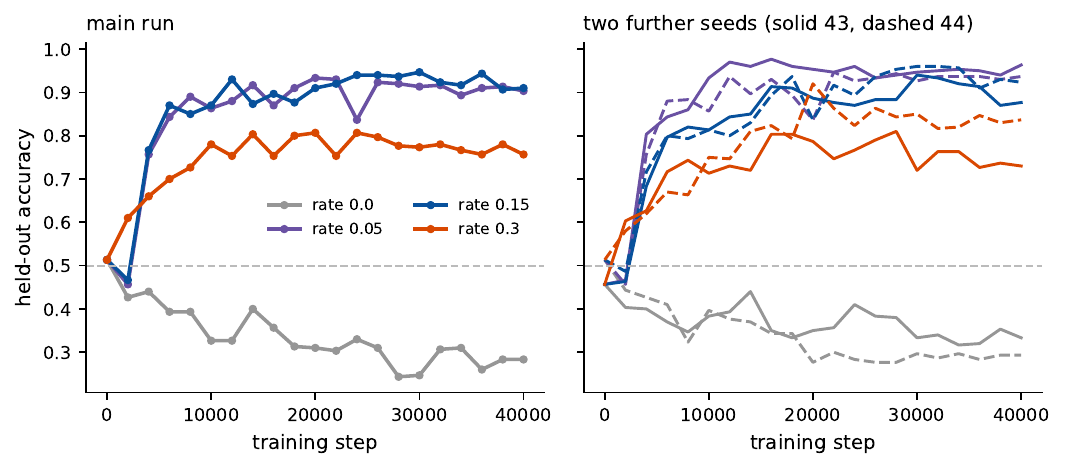}
\caption{Held-out accuracy for the four injection rates. The left panel shows the main run and the right panel two further seeds, with seed 43 solid and seed 44 dashed.}
\label{fig:appctrl}
\end{figure*}

At injection rate $0.05$ held-out accuracy falls to $0.457$ at step 2000 before recovering
to $0.903$, with in-distribution accuracy reaching $1.000$. At rate $0.15$ the principal run
reaches a minimum of $0.467$ and recovers to $0.910$, again reaching $1.000$ in
distribution. At rate $0.30$ held-out accuracy does not fall below its starting value and
ends at $0.757$, while in-distribution accuracy reaches $1.000$. At rates $0.05$ and $0.15$ the temporary
below-chance point followed by recovery appears across all three training seeds, whereas
the zero-injection models remain below chance and the $0.30$ condition is already above
chance by the first post-initial evaluation. Figure~\ref{fig:appctrl} shows all three
trajectories for each rate.

\begin{figure}[t]
\centering
\includegraphics[width=\columnwidth]{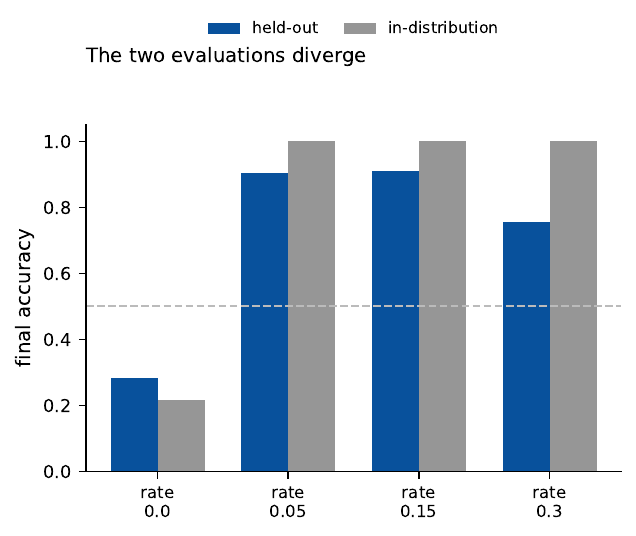}
\caption{Final accuracy on the two evaluations for each injection rate in the main run.}
\label{fig:appctrldist}
\end{figure}

Intermediate injection rates therefore produce both the clearest temporary deterioration
and the strongest held-out generalisation. Zero injection produces no task learning, while
the highest rate permits ceiling in-distribution performance but weaker transfer to unseen
names. Figure~\ref{fig:appctrldist} plots both evaluations. The relationship between injection rate and held-out behaviour is non-monotonic.

Within this training setup, the temporary dip depends on the frequency of task-structured examples and is not an unavoidable consequence of eventually learning the task. The intermediate injection rates produce the clearest temporary deterioration and the strongest held-out generalisation.

\subsection{A minimal symbolic task}
\subsubsection{Task}
We construct a minimal task that retains the competition between ordinary copying and a
rarer override rule while removing natural-language structure. Each sequence contains three
symbols drawn from a vocabulary of size $V$. An ordinary copy example contains three
distinct symbols and targets the final symbol,
\[
[x,\, y,\, z] \rightarrow z ,
\]
while an override example repeats the first symbol in third position but targets the middle
symbol,
\[
[a,\, b,\, a] \rightarrow b .
\]
A model that has learned only the ordinary copying rule succeeds on copy examples but
systematically predicts the repeated final symbol on override examples. This is a minimal
analogue of the competition studied in the rest of the paper, not the task itself. Here
copying refers only to the behavioural rule of predicting the final input symbol. It is not
intended to reproduce the attention-head copying mechanism identified in the Pythia
models. Chance
accuracy on override examples is $1/V$ rather than one half, because the prediction is made
over the full candidate-symbol vocabulary.

\subsubsection{Models and training sweep}
The models are two-layer transformers with four attention heads. We vary the model
dimension over $\{16, 32, 64\}$ with the feed-forward width fixed at twice that, the
vocabulary size over $\{50, 100\}$, the learning rate over
$\{3\times10^{-5}, 10^{-4}, 3\times10^{-4}\}$, the training length over 30000 or 50000
steps, and the copy-example fraction over $\{0.90, 0.95, 0.98\}$. The remainder of each
batch consists of override examples, and we refer to the copy-example fraction as copy
pressure. The large number of possible symbol combinations prevents the task from reducing
to memorisation of a small fixed list of override triples.

\subsubsection{Intervention and criteria}
At each evaluated checkpoint we substitute the token at the repeated position. For every
example a replacement symbol is drawn uniformly from the vocabulary excluding both the
repeated symbol and the override target, and its embedding, together with the position
embedding for that slot, replaces the original. The substitution is therefore an
independent per-example change of symbol identity at the input embedding rather than a
random continuous vector, and no norm matching is applied. We then measure the change in
the logit difference between the override target and the repeated symbol.

A positive effect means that disrupting the repeated position improves the preference for
the override target, and a negative effect means the same disruption harms it.

We apply two trajectory criteria. A behavioural dip occurs when the lowest
evaluated override accuracy after the initial checkpoint falls below chance. An intervention
reversal occurs when the measured effect is above $+0.1$ at one evaluated checkpoint and
below $-0.1$ at the next evaluated checkpoint, which is how the criterion is implemented.

\subsubsection{Results}
Ten runs are evaluated. Eight vary vocabulary size, model dimension, learning rate, training
duration, and copy pressure. Two further runs repeat the strongest configuration under new
seeds, giving three seeds for that configuration in total.

\begin{figure}[t]
\centering
\includegraphics[width=\columnwidth]{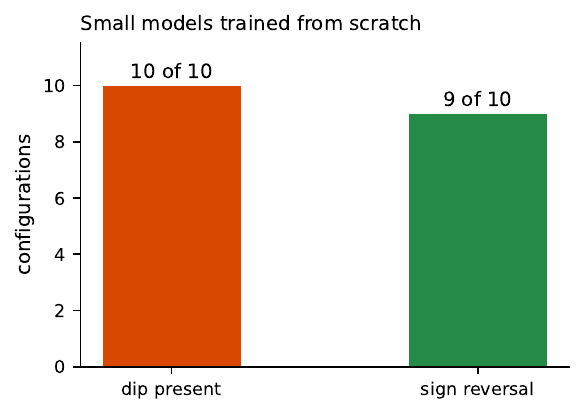}
\caption{Ten runs of the minimal task trained from scratch. All ten show the dip and nine show the sign reversal.}
\label{fig:appmini}
\end{figure}

All ten runs exhibit a behavioural dip, and nine of the ten exhibit the intervention
reversal. Figure~\ref{fig:appmini} counts the runs meeting each criterion. Together, these
runs provide a controlled sufficiency result: a training distribution dominated by an
ordinary copying rule but containing a rarer override rule can produce both temporary
below-chance behaviour and a reversal in the causal role of the repeated-position state.
The experiment establishes this result at the level of the training distribution, while its
internal mechanism and relationship to natural-language pretraining remain open.

\subsection{Models trained only on ordinary text}
\subsubsection{Setup}
The controlled-injection experiments deliberately add task-structured sequences. We
therefore perform a separate test using two small language models trained by the authors
on ordinary web text with no task-specific sequences added.

Both are GPT-2 architecture models with eight layers, model dimension 512, and eight
attention heads, using the GPT-2 tokenizer and its vocabulary of 50257, trained on a
stream of OpenWebText at learning rate $6\times10^{-4}$. The two runs differ in length: the
first was trained to step 200000 and yields 65 evaluated checkpoints, the second to step
105000 and yields 46.

At every checkpoint we evaluate the same repeated-name task on a fixed set of 200 prompts.
We also construct a matched control prompt in which the second occurrence of the repeated
name is replaced by a third single-token name. The evaluated prompt itself remains
unchanged. Instead, across an early-to-middle band of transformer blocks running from one
quarter to one half of the depth, we replace its hidden state at the second mention with
the corresponding state produced by the control prompt, and measure the resulting change in
logit difference. This hidden-state replacement leaves the evaluated tokens unchanged and intervenes only on the internal state at the second mention.

\subsubsection{Behavioural trajectory}
Both models remain close to chance in the two-name comparison, but their mean logit
difference becomes negative over substantial portions of training. The first run reaches
$-0.260$ at step 25000 and the second reaches $-0.259$ at step 40000. The first ends at
$+0.141$ and the second at $-0.231$, so neither develops the strong recovered preference
seen in mature Pythia.

\begin{figure*}[t]
\centering
\includegraphics[width=\textwidth]{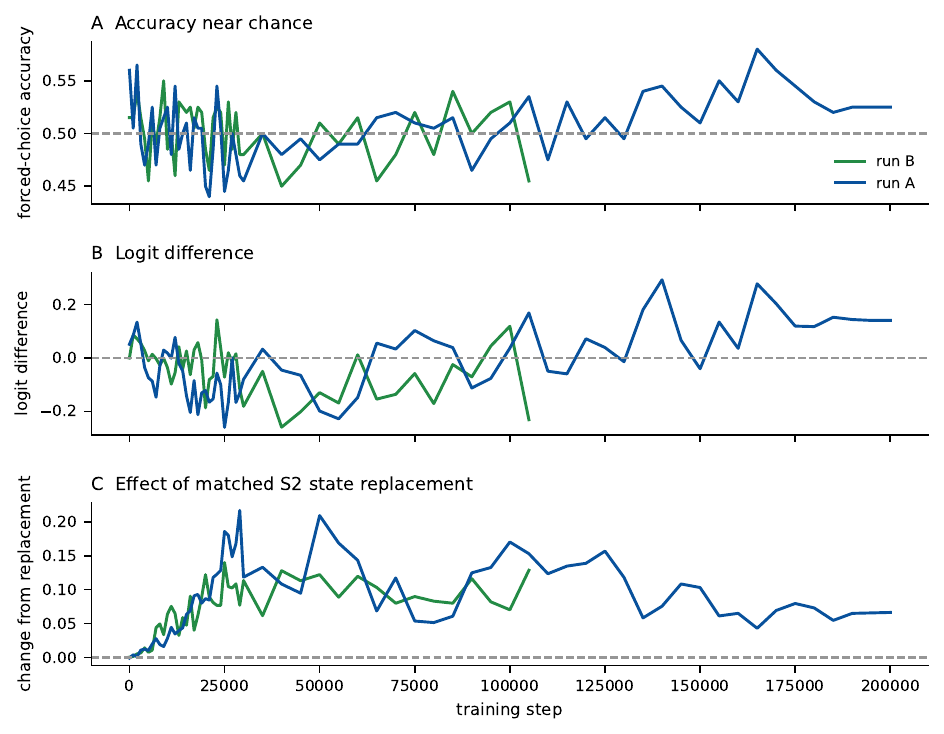}
\caption{Two models trained on ordinary web text. Panel A shows forced-choice accuracy, panel B the logit difference, and panel C the change produced by replacing the residual-stream state at the second occurrence of the repeated name with the corresponding state from a matched control prompt. The evaluated prompt tokens are unchanged. Run A covers 200000 steps and run B covers 105000 steps.}
\label{fig:appowt}
\end{figure*}

The replacement improves the logit difference at 64 of the 65 evaluated checkpoints in the
first run and at all 46 in the second. This establishes that a repetition-sensitive
preference against the correct name can emerge during ordinary-text training without
explicit injection of the evaluation task. Figure~\ref{fig:appowt} plots both runs.

These runs show that harmful repetition sensitivity can emerge during ordinary-text training without explicit task injection. Because neither model develops a strong recovered preference, they capture the harmful early side of the pattern without its later corrective phase.

\subsection{Complementary evidence from three training settings}
The three experiments provide complementary evidence.

Controlled task injection shows that the presence and depth of temporary deterioration
depend on the frequency of task-structured examples, and that ceiling performance on
familiar names does not imply equivalent generalisation to unseen names.

The minimal symbolic task gives a sufficiency result. A common copying rule combined with a
rarer override rule can produce both a behavioural dip and a causal reversal in models
trained from scratch.

The ordinary-text runs show that harmful repetition sensitivity can emerge without explicit
task injection. Because those models never reach a strong recovered state, they do not
establish that ordinary web-text training alone reproduces the complete copying-to-suppression trajectory.

These experiments make the developmental phenomenon easier to reproduce and constrain
possible explanations for it. Determining why the natural Pythia training distribution learns harmful copying before mature suppression remains an open training-dynamics question.

\section{Earlier Version and Reproducibility}
\label{app:repro}

\subsection{Relationship to the Workshop Version}

A preliminary version of this work reported the initial training-time sign flip \citep{dahiya2026signflip}. This paper develops the causal account and expands the evaluation across interventions, controls, model scales, model families, and training settings.

\subsection{Released Artefacts}

The code and results supporting this paper are available at \url{https://github.com/Tejas7007/EMNLP_Findings_2026_Copying_Before_Suppression}, and the separately trained Pythia-160M model with its dense checkpoints through the early training transition is available at \url{https://huggingface.co/teys7007/pythia-160m-seed42-dense} (DOI: \texttt{10.57967/hf/10155}).

\end{document}